\documentclass[acmsmall, nonacm=true]{acmart}

\usepackage{mathtools, nccmath}
\DeclarePairedDelimiter{\nint}\lfloor\rceil

\makeatletter
\newcommand{\thickhline}{%
    \noalign {\ifnum 0=`}\fi \hrule height 0.65pt
    \futurelet \reserved@a \@xhline
}
\makeatother

\usepackage{bm}
\usepackage{physics}
\usepackage{multirow}
\usepackage[symbols,nogroupskip,sort=none]{glossaries-extra}

\newcommand{\eloss}{$S$-Loss}
\newcommand{\nsloss}{$C$-Loss{}} 
\newcommand{\mse}{$MSE${}}

\newcommand{\proposed}{\emph{CoPhy}-PGNN}
\newcommand{\nexmodel}{\emph{CoPhy}-PGNN (only-$\mathcal{D}_{Tr}$)}
\newcommand{\nemodel}{\emph{CoPhy}-PGNN (w/o \eloss)}

\newcommand{\lfmodel}{\emph{CoPhy}-PGNN (Label-free)}
\newcommand{\ncmodel}{PGNN-\emph{analogue}}
\newcommand{\nn}{Black-box NN}
\newcommand{\pinn}{PINN-\emph{analogue}}
\newcommand{\pgnn}{PGNN-\emph{analogue}}
\newcommand{\vpgnn}{\emph{MTL}-PGNN}
\newcommand{\eigenvec}{\boldsymbol{y}}
\newcommand{\eigenval}{b}
\newcommand{\eigenmat}{\hat{A}}

\glsxtrnewsymbol[description={Sets the asymptotic value of $\lambda_{C}$ after a sufficient number of epochs.}]{a}{\ensuremath{\lambda_{C0}}}
\glsxtrnewsymbol[description={Sets the rate of growth of $\lambda_{C}$.}]{b}{\ensuremath{\alpha_C}}
\glsxtrnewsymbol[description={Sets the cut-off number of epochs after which $\lambda_C$ is activated from a cold start of 0.}]{c}{\ensuremath{T_{a}}}
\glsxtrnewsymbol[description={Sets the starting value of $\lambda_{S}$.}]{d}{\ensuremath{\lambda_{S0}}}
\glsxtrnewsymbol[description={Sets the rate of annealing of $\lambda_{S}$.}]{e}{\ensuremath{\alpha_S}}
\glsxtrnewsymbol[description={Sets the frequency of the annealing update for $\lambda_{S}$.}]{f}{\ensuremath{T}}

\AtBeginDocument{%
  \providecommand\BibTeX{{%
    \normalfont B\kern-0.5em{\scshape i\kern-0.25em b}\kern-0.8em\TeX}}}

\setcopyright{none}
\copyrightyear{2018}
\acmYear{2018}
\acmDOI{10.1145/1122445.1122456}

\acmJournal{JACM}
\acmVolume{37}
\acmNumber{4}
\acmArticle{}
\acmMonth{8}

\begin{document}

\title[\proposed{}: Learning Physics-guided Neural Networks with Competing Loss Functions]{\proposed{}: Learning Physics-guided Neural Networks with Competing Loss Functions for Solving Eigenvalue Problems}

\author{Mohannad Elhamod}
\authornote{Both authors contributed equally to this research.}
\email{elhamod@vt.edu}
\orcid{0000-0002-2383-947X}
\author{Jie Bu}
\authornotemark[1]
\email{jayroxis@vt.edu}
\affiliation{%
  \institution{Department of Computer Science, Virginia Tech}
  \city{Blacksburg}
  \state{Virginia}
  \country{USA}
  \postcode{24061}
}

\author{Christopher Singh}
\affiliation{%
  \institution{Department of Physics, Binghamton University}
  \city{Binghamton}
  \state{New York}
  \country{USA}}
\email{csingh5@binghamton.edu}

\author{Matthew Redell}
\affiliation{%
  \institution{Department of Physics, Binghamton University}
  \city{Binghamton}
  \state{New York}
  \country{USA}
}
\email{mredell1@binghamton.edu}

\author{Abantika Ghosh}
\affiliation{%
 \institution{Department of Physics and Applied Physics, University of Massachusetts Lowell}
 \city{Lowell}
 \state{Massachusetts}
 \country{USA}}
\email{Abantika_Ghosh@student.uml.edu}

\author{Viktor Podolskiy}
\affiliation{%
  \institution{Department of Physics and Applied Physics, University of Massachusetts Lowell}
 \city{Lowell}
 \state{Massachusetts}
 \country{USA}}
\email{Viktor_Podolskiy@uml.edu }

\author{Wei-Cheng Lee}
\affiliation{%
  \institution{Department of Physics, Binghamton University}
  \city{Binghamton}
  \state{New York}
  \country{USA}
}
\email{wlee@binghamton.edu}

\author{Anuj Karpatne}
\affiliation{%
  \institution{Department of Computer Science, Virginia Tech}
  \city{Blacksburg}
  \state{Virginia}
  \country{USA}
  \postcode{24061}
}
\email{karpatne@vt.edu}

\renewcommand{\shortauthors}{Elhamod and Bu, et al.}

\begin{abstract}
Physics-guided Neural Networks (PGNNs) represent an emerging class of neural networks that are trained using physics-guided (PG) loss functions (capturing violations in network outputs with known physics), along with the supervision contained in data. Existing work in PGNNs has demonstrated the efficacy of adding single PG loss functions in the neural network objectives, using constant trade-off parameters, to ensure better generalizability. However, in the presence of multiple PG functions with competing gradient directions, there is a need to \textit{adaptively} tune the contribution of different PG loss functions during the course of training to arrive at generalizable solutions. We demonstrate the presence of competing PG losses in the generic neural network problem of solving for the lowest (or highest) eigenvector of a physics-based eigenvalue equation, which is commonly encountered in many scientific problems. We present a novel approach to handle competing PG losses and demonstrate its efficacy in learning generalizable solutions in two motivating applications of quantum mechanics and electromagnetic propagation.
All the code and data used in this work are available at \url{https://github.com/jayroxis/Cophy-PGNN}.
\end{abstract}





\maketitle

\section{Introduction}

With the increasing impact of deep learning methods in diverse scientific disciplines \cite{Appenzeller16,graham2008big}, there is a growing realization in the scientific community to harness the power of artificial neural networks (ANNs) without ignoring the rich supervision available in the form of physics knowledge in several scientific problems \cite{karpatne2017theory,willard2020integrating}. 
One of the promising lines of research in this direction is to modify the objective function of neural networks by adding loss functions that measure the violations of ANN outputs with physical equations, termed as \textit{physics-guided (PG) loss functions} \cite{karpatne2017theory,stewart2017label}. 
By anchoring ANN models to be {consistent} with physics, PG loss functions have been shown to  impart generalizability even in the paucity of training data across several scientific problems
\cite{jia2019physics,karpatne2017physics,raissi2019physics,de2019deep}. We refer to the class of neural networks that are trained using PG loss functions as physics-guided neural networks (PGNNs).

\begin{figure}[t]
\includegraphics[width=0.44\textwidth]{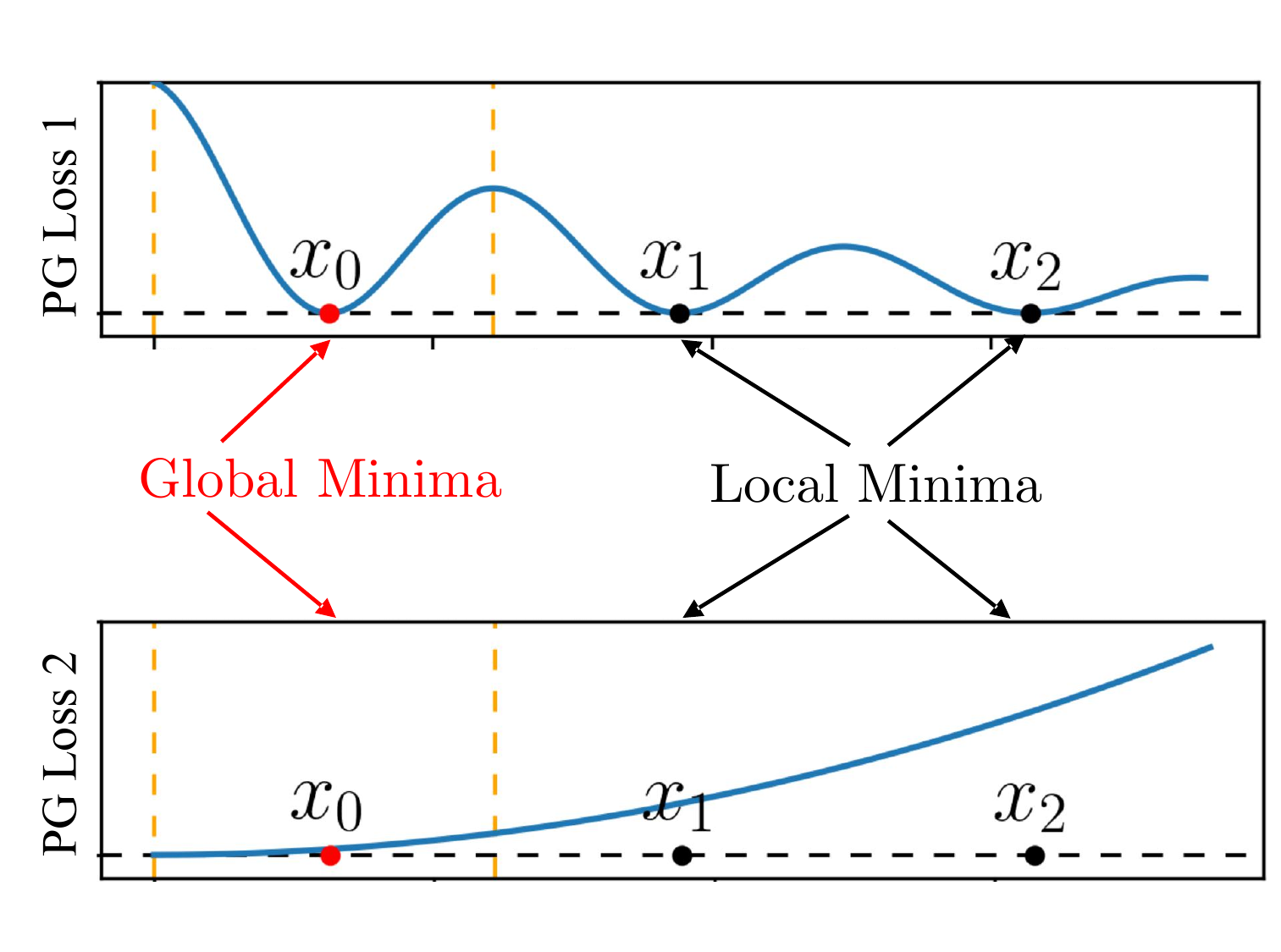}
\captionsetup{font=small}
\caption{A toy example showing competing Physics-guided (PG) losses that can lead to poor local minima when minimized together. The $Y$ axis denotes the value of the loss terms and the $X$ axis denotes the space of model parameters.}
\label{fig:example}
\end{figure}

While some existing work in PGNN have attempted to learn neural networks by solely minimizing PG loss (and thus being label-free) \cite{raissi2019physics,stewart2017label}, others have used both PG loss and data-driven loss (measured on a labeled set) using appropriate trade-off hyper-parameters \cite{karpatne2017physics,jia2019physics}. Note that finding the optimal setting of these hyper-parameters can be challenging because of the varying nature and gradient dynamics of PG loss and data-driven loss terms \cite{wang2020understanding,wang2020and}.
An even more challenging problem is when there are multiple physics equations with \emph{competing} PG loss functions that need to be minimized together, where each PG loss may show multiple local minima. In such situations, simple addition of PG losses in the objective function with constant trade-off hyper-parameters may result in the learning of non-generalizable solutions. 


Figure \ref{fig:example} shows a toy example with two PG loss terms. We use this example to demonstrate how optimizing multiple PG loss terms with varying properties (e.g., curvatures) can lead to learning inferior solutions because of \emph{competition} between their gradients. In particular, if we add the two loss terms with constant weighting coefficients and optimize the weighted sum using gradient descent methods, it is possible to end up at a local minimum of both loss functions (e.g. $\mathbf{x_1}$ or $\mathbf{x_2}$) if we start from an inferior setting of the model parameters (e.g., an initial value closer to $\mathbf{x_1}$ or $\mathbf{x_2}$). However, if we only pay importance to PG loss 2 (which varies smoothly and does not suffer from many local minima) in the initial epochs of gradient descent, we will be able to arrive at a solution that is closer to the global minimum, i.e. $\mathbf{x_0}$. In the later epochs, we can then optimize PG loss 1 to arrive at the global minimum $\mathbf{x_0}$. Note that this strategy would arrive at a generalizable solution regardless of where the optimization process started in the parameter space. This observation, although derived from a simple toy problem, motivates us to ask the question: \emph{is it possible to adaptively balance the importance of competing PG loss functions at different stages of neural network learning to arrive at generalizable solutions?}


In this work, we introduce our proposed framework of \proposed{}, which is an abbreviation for \emph{\underline{Co}mpeting \underline{Phy}sics} \underline{P}hysics-\underline{G}uided \underline{N}eural \underline{N}etworks, to handle the learning of multiple PG loss functions during neural network training. Our proposed framework derives inspiration from continuation methods \cite{bengio2009curriculum} used in the optimization literature for solving non-convex problems, where the strategy is to focus on ``simpler'' components of the objective function before getting to the more ``complex'' ones. While the basic idea of our approach is applicable across many problems, we specifically develop our framework for the problem of solving eigenvalue problems that are encountered in many scientific disciplines. In particular, we target scientific problems where physics knowledge is represented as eigenvalue equations and we are required to solve for the highest or lowest eigen-solution. This representation is common to many types of physics knowledge such as the Schr\"{o}dinger equation in the domain of quantum mechanics and Maxwell's equations in the domain of electromagnetic propagation. In these applications, solving eigenvalue equations using exact numerical techniques (e.g., diagonalization methods) can be computationally expensive especially for large physical systems. On the other hand, PGNN models, once trained, can be applied on testing scenarios to predict their eigen-solutions
in drastically smaller running times. We empirically demonstrate the efficacy of our \proposed{} solution on two diverse applications in quantum mechanics and electromagnetic propagation, highlighting the generalizability of our proposed approach to many physics problems.

The remainder of the paper is organized as follows. Section \ref{sec:bkg} provides background of our physics problems and describes related work. Section \ref{sec:method} presents our proposed approach. Section \ref{sec:eval} describes evaluation setup. Section \ref{sec:results} presents our results. Finally, Section \ref{sec:conclusion} concluding remarks, respectively.

\section{Background}
\label{sec:bkg}

\subsection{Overview of Physics Problems}


While our target applications of quantum mechanics and electromagnetic propagation represent two distinct domains, they share a common property that the physics of the problem is available in the form of an eigenvalue equation, which is common to many scientific problems. In particular, the physics of both problems can be represented as
$\eigenmat{}\eigenvec{} = \eigenval{}\eigenvec{}$, where $\eigenmat{}$ is the input matrix, $\eigenval{}$ is an eigenvalue of the input matrix and $\eigenvec{}$ is the corresponding eigenvector. We are interested in solving the lowest or highest eigen-solution of this equation in our target problems. We provide a brief overview of both target applications in the following (detailed overviews are provided in
Appendix \ref{sec:background}).

\paragraph{Quantum Mechanics:}
In this application, the goal is to predict the ground-state wave function of an Ising chain model with $n = 4$ particles. The physics of this problem can be described by the Schr\"{o}dinger equation, which can be stated as  $\mathbf{H}\hat{\Psi}=\hat{E} \hat{\Psi}$, where $\hat{E}$ is the energy level of the model (or the eigenvalue of the equation), $\hat{\Psi}$ is the wave function (or the eigenvector), and $\mathbf{H}$ is the Hamiltonian  matrix describing the input conditions of the Ising chain model. Since the ground-state wave function corresponds to the lowest energy level, we are interested in finding the lowest eigen-solution of this eigenvalue equation. 

\paragraph{Electromagnetic Propagation:}
Our second application involves the propagation of electromagnetic waves in periodically stratified layer stacks. The description of this propagation can be reduced to the eigenvalue problem $\hat A \vec{h_m}=k_z^{m^2} \vec{h_m}$,  where $k_z^{m^2}$, the propagation constant of the electromagnetic modes along the layers, is the eigenvalue; and $\vec h_m$, the coefficients of the Fourier transform of the spatial profile of the electromagnetic field, is the eigenvector. It is important to note that these quantities are complex valued, unlike in the target application of quantum mechanics. Also, unlike the quantum mechanics problem, we are interested in the largest eigenvalue rather than the smallest. {Another basic difference is the large size of input matrices we considered in this problem in contrast to the quantum mechanics problem to test the scalability of our method to large systems.}

\subsection{Related Work in PGNN}

PGNN has found successful applications in several disciplines including fluid dynamics \cite{wang2017physics,wang2017comprehensive}, climate science \cite{de2019deep}, and lake modeling \cite{karpatne2017physics,jia2019physics,daw2020physics}. However, to the best of our knowledge, PGNN formulations have not been explored yet for our target applications of solving eigenvalue equations in  the field of quantum mechanics and electromagnetic propagation. Existing work in PGNN can be broadly divided into two categories. The first category involves label-free learning by only minimizing PG loss without using any labeled data.
For example, physics-informed neural networks (PINNs) and its variants \cite{raissi2019physics,raissi2017physics1,raissi2017physics2} have been recently developed to solve PDEs by solely minimizing PG loss functions, for simple canonical problems such as Burger's equation. Since these methods are label-free, they do not explore the interplay between PG loss and label loss. We consider an analogue of PINN for our target application as a baseline in our experiments.

The second category of methods incorporate PG loss as additional loss terms in the objective function along with label loss, using constant trade-off hyper-parameters. This includes work in basic Physics-guided Neural Networks (PGNNs)  \cite{karpatne2017physics,jia2019physics} for the target application of lake temperature modeling. We use an analogue of this basic PGNN as a baseline in our experiments.

While some recent works  have investigated the effects of PG loss on generalization performance \cite{shin2020convergence} and the importance of normalizing the scale of hyper-parameters corresponding to PG loss terms \cite{wang2020understanding}, they do not study the effects of competing PG losses which is the focus of this paper. In particular, our empirical study reveals that different loss terms have different importance at different stages of learning, and it is important to weight them differently (rather than balancing their contributions) using adaptive weighting coefficients to ensure better generalizability. Our work is related to the field of multi-task learning (MTL) \cite{Multitask,chen2018gradnorm,sener2019multitask}, as the minimization of PG losses and label loss can be viewed as multiple shared tasks. For example, alternating minimization techniques in MTL \cite{AlternativeOptimization} can be used to alternate between minimizing different PG loss and label loss terms over different mini-batches. We consider this as a baseline approach in our experiments. 
{ Also note that the use of alternating mechanisms in neural networks guided by physics knowledge have been developed for inverse problems, e.g., in a recent work for Fourier ptychography (FP) \cite{zhang2019pgnn}. However, this work is different from the focus of this paper on solving eigenvalue problems using physics knowledge and data.
}

\section{Methodology}
\label{sec:method}

\subsection{Problem Statement}
From an ML perspective, we are given a collection of training pairs, $\mathcal{D}_{Tr}:=\{\eigenmat{}_i,(\eigenvec_i,\eigenval_i)\}_{i=1}^N$, where $(\eigenvec_i,\eigenval_i)$ is generated by diagonalization solvers (considered as expert ground-truth). We consider the problem of learning an ANN model, $(\hat{\eigenvec},\hat{\eigenval}) = f_{NN}(\eigenmat,\mathbf{\theta})$, that can predict $(\eigenvec,\eigenval)$ for any input matrix, $\eigenmat$, where $\mathbf{\theta}$ are the learnable parameters of ANN. We are also given a set of unlabeled examples, $\mathcal{D}_{U}:=\{\eigenmat_i\}_{i=1}^M$, which will be used for testing. We consider a simple feed-forward architecture of $f_{NN}$ in all our formulations. 

\subsection{Designing Physics-guided (PG) Loss Functions} 

A na\"ive approach for learning $f_{NN}$ is to minimize the mean sum of squared errors (MSE) of predictions on the training set, referred to as the $\text{Train-MSE}(\theta):=1/N (\sum_{i}||\hat{\eigenvec}_i - \eigenvec_i||^2 + ||\hat{\eigenval}_i - \eigenval_i||^2$). However, instead of solely relying on Train-MSE, we consider the following PG loss terms to guide the learning of $f_{NN}$ to generalizable solutions:

\paragraph{Characteristic Loss:} 
A fundamental equation we want to satisfy in our predictions, $(\hat{\eigenvec},\hat{\eigenval})$, for any input $\eigenmat$ is the eigenvalue equation, $\eigenmat \hat{\eigenvec} = \hat{\eigenval} \hat{\eigenvec}$. Hence, we consider minimizing the following equation:
\begin{equation}
    \text{\nsloss{}}(\theta):= \frac{1}{N} \sum_i \frac{||\eigenmat_i \hat{\eigenvec}_i - \hat{\eigenval}_i \hat{\eigenvec}_i||^2}{\hat{\eigenvec}^{\top}\hat{\eigenvec}},
\end{equation}
where the denominator term ensures that $\hat{\eigenvec}$ does not tend toward the trivial solution of 0. By construction, \nsloss{} only depends on the predictions of $f_{NN}$ and does not rely on true labels, $(\eigenvec,\eigenval)$. Hence, it can be evaluated even on the unlabeled test data, $\mathcal{D}_{U}$. 

\paragraph{Spectrum Loss:}
\label{subsec:eloss}
Note that every solution of the equation  $\eigenmat \hat{\eigenvec} = \hat{\eigenval} \hat{\eigenvec}$ is a ``local minimum'' in the optimization landscape of \nsloss{}. 
In particular, for every input $\eigenmat_i \in \mathcal{D}_{U}$, there are $d$ possible eigen-solutions (where $d$ is the length of $\hat{\eigenvec}$), each of which will result in a perfectly low value of \nsloss{} $=0$, thus acting as a local minimum.
However, we are only interested in a specific eigen-solution for every $\eigenmat_i$, namely the eigen-solution that has the smallest or the largest eigenvalue. Therefore, we consider minimizing another PG loss term that ensures the predicted $\hat{\eigenval}$ at every sample is optimal. In the case of the quantum mechanics application, we use the following loss to find the smallest eigen-solution:
\begin{equation}
    \text{\eloss{}}(\theta):=\frac{1}{N}\sum_i \exp(\hat{\eigenval}_i)
\end{equation}
The use of $\exp$ function ensures that E-Loss is always positive, even when predicted eigenvalues are negative.
For the electromagnetic propagation application, we simply direct the optimization towards the largest eigenvalue by replacing $\hat{\eigenval}_i$ with $-\Re(\hat{\eigenval}_i)$, where $\Re$ extracts the real part of the complex eigenvalue. This can be generalized to other problems where we want to target a specific eigen-solution for every input matrix.

\subsection{Proposed Approach for Adaptively Tuning PG Losses}

{
\paragraph{Rationale for Proposed Approach:} 
PG loss terms can be integrated in the learning objective of $f_{NN}$ by adding them to Train-MSE using trade-off  coefficients $\lambda_C$ and $\lambda_S$ for \nsloss{} and \eloss{}, respectively. Conventionally, such trade-off coefficients are kept constant across all epochs of gradient descent. This inherently assumes that the importance of PG loss terms in guiding the learning of $f_{NN}$ towards a generalizable solution is constant across all stages (or epochs) of gradient descent, and they are in agreement with each other.

However, in many scientific problems, PG loss terms may show \textit{competing} directions of gradient descent because of differences in their properties (e.g., curvatures), making it difficult to balance them at different stages of ANN training. For example, in the problem of solving eigenvalue equations, we can see that the two PG loss terms, \nsloss{} and \eloss{}, have widely different number of local minima and thus have different curvatures. On one hand, \nsloss{} suffers from a large number of local minima since every eigen-solution of  $\eigenmat_i$ has $\text{\nsloss{}} = 0$. Among all these solutions, we are only interested in eigen-solutions with the smallest (or largest) eigenvalues across all $\eigenmat_i$. On the other hand, \eloss{} is a monotonic function and thus has a single minimum in a bounded domain. This behavior of \nsloss{} and \eloss{} is similar to the competition among PG loss terms illustrated in Figure \ref{fig:example}, where a simple combination of PG loss terms can easily lead to poor local minima solutions. 
We thus need a way to \textit{adaptively tune} the importance of competing PG loss terms at different training epochs to steer the learning dynamics away from poor local minima and  toward basins of attraction of dominant (if not global) minima. 

To this effect, we present our proposed framework of \proposed{}, which is an abbreviation for \emph{\underline{Co}mpeting \underline{Phy}sics}  PGNN. The basic idea of our proposed framework is to prefer PG loss terms that show fewer local minima (e.g., \eloss) to dominate the gradient descent updates in the beginning epochs of ANN training, when the network is susceptible to making larger jumps and thus getting stuck at poor local minima. Once we have crossed a sufficient number of epochs and are already close to a generalizable solution $\theta^*$, we can let more complex PG loss terms with larger number of local minima (e.g., \nsloss{}) dominate the gradient descent dynamics and refine the learned parameters to converge to $\theta^*$. Our approach derives inspiration from \textit{continuation methods} developed in  the scientific optimization literature for non-convex problems \cite{allgower2003introduction}, where we first optimize a simpler (smoothed) objective function and gradually consider less smoothing. It is also related to work in the area of {curriculum learning} \cite{bengio2009curriculum,elman1993learning,soviany2021curriculum}, where ``simpler'' examples are presented to the neural network in an incremental manner before showing more ``complex'' ones.

In the following, we present the two primary components of our \proposed{} framework: (i) annealing $\lambda_S$, the trade-off coefficient of \eloss{}, to start with a large value that gradually decays over epochs, and (ii) cold starting $\lambda_C$, the trade-off coefficient for \nsloss{}, to take non-zero values only in the later epochs. A list of all hyper-parameters used in \proposed{} is presented as a Table at the end of the Appendix.


}

\paragraph{Annealing $\lambda_S$:} 
We want \eloss{} to dominate the training dynamics in the beginning epochs because \eloss{} does not suffer from multiple local minima w.r.t. outputs and hence has a smooth curvature. Hence, we perform simulated annealing of $\lambda_S$ such that it takes on a high value in the beginning epochs and slowly reduces to 0 in later epochs. Specifically, we consider the following annealing procedure for $\lambda_S$:
\begin{equation}
    \lambda_S(t) = \lambda_{S0} \times (1 - \alpha_S)^{\nint{t/T}}
\end{equation}
where, $\lambda_{S0}$ is a hyper-parameter denoting the starting value of $\lambda_{S}$ at epoch 0,  $\alpha_S < 1$ is a hyper-parameter that controls the rate of annealing, and $T$ sets the frequency of the annealing update.

\paragraph{Cold Starting $\lambda_C$:} 

Since the \nsloss{} is highly non-convex w.r.t outputs and suffers from a large number of local minima, we want to turn off its importance in the beginning epochs to avoid getting stuck at poor local minima. 
We thus perform ``cold starting'' of $\lambda_C$, where its value is kept to 0 in the beginning epochs after which it is raised to a constant value, as given by the following procedure:
\begin{equation}
    \lambda_C(t) = \lambda_{C0} \times sigmoid(\alpha_C \times (t - T_{a}))
\end{equation}
where, $\lambda_{C0}$ is a hyper-parameter denoting the final value of $\lambda_{C}$ after a sufficient number of epochs,  $\alpha_C$ is a hyper-parameter that dictates the rate of growth of the sigmoid function, and $T_{a}$ is a hyper-parameter that controls the cut-off number of epochs after which $\lambda_C$ is activated from a cold start of 0.

\paragraph{Overall Learning Objective of \proposed{}:}
After incorporating the adaptive tuning strategies of $\lambda_S$ and $\lambda_C$, the overall learning objective of \proposed{} can be written as:
\begin{equation}
    E(t) = \text{Train-Loss} + \lambda_C(t) ~ \text{\nsloss{}} + \lambda_S(t)  ~\text{\eloss{}} \nonumber
\end{equation}
Note that Train-Loss is only computed over  $\mathcal{D}_{Tr}$, whereas the PG loss terms, \nsloss{} and \eloss{}, are computed over $\mathcal{D}_{Tr}$ as well as the set of unlabeled samples, $\mathcal{D}_{U}$. 
\section{Evaluation setup}
\label{sec:eval}

\paragraph{Data in Quantum Mechanics:}


We considered $n=4$ spin systems of Ising chain models for predicting their ground-state wave-function under varying influences of two controlling parameters: $B_x$ and $B_z$, which represent the strength of external magnetic field along the $X$ axis (parallel to the direction of Ising chain), and $Z$ axis (perpendicular to the direction of the Ising chain), respectively. The Hamiltonian matrix $\mathbf{H}$ for these systems is then given as:

\begin{equation}
	\mathbf{H} = - \sum_{i=0}^{n-1} \sigma_i^z \sigma_{i+1}^z 
	- B_x \sum_{i=0}^{n-1} \sigma_i^x 
	- B_z \sum_{i=0}^{n-1} \sigma_i^z,
\end{equation}

where $\sigma^{x,y,z}$ are Pauli operators and ring boundary conditions are imposed. Note that the size of $\mathbf{H}$ is $d = 2^n = 16$. We set $B_z$ to be equal to 0.01 to break the ground state degeneracy, while $B_x$ was sampled from a uniform distribution from the interval [0, 2]. 

Note that when $B_x < 1$, the system is said to be in a ferromagnetic phase, since all the spins prefer to either point upward or downward collectively. However, when $B_x > 1$, the system transitions to paramagnetic phase, where both upward and downward spins are equally possible. Because the ground-state wave-function behaves differently in the two regions, the system actually exhibits different physical properties. Hence, in order to test for the generalizability of ANN models when training and test distributions are different, we generate training data only from the region deep inside the ferromagnetic phase for $B_x<0.5$, while the test data is generated from a much wider range $0<B_x<2$, covering both ferromagnetic and paramagnetic phases. In particular, the training set comprises of $N=100,000$ points with $B_x$ uniformly sampled from $0$ to $0.5$, while the test set comprises of $M=20,000$ points with $B_x$ uniformly sampled from $0$ to $2$. Labels for the ground-state wave-function for all training and test points were obtained by  direct diagonalization of the Ising Hamiltonian using Intel's implementation of LAPACK (MKL). We used uniform sub-sampling and varied $N$ from $100$ to $20,000$ to study the effect of training size on the generalization performance of comparative ANN models. For validation, we also used sub-sampling on the training set to obtain a validation set of $2000$ samples. We performed 10 random runs of uniform sampling for every value of $N$, to show the mean and variance of the performance metrics of comparative ANN models, where at every run, a different random initializtion of the ANN models is also used. Unless otherwise stated, the results in any experiment are presented over training size $N = 2000$. 


\paragraph{Data in Electromagnetic Propagation:}
We considered a periodically stratified layer stack of 10 layers of equal length per period. The refractive index $n$ of each layer was randomly assigned an integer value between 1 and 4. Hence, the permittivity $\epsilon=n^2$ can take values from $\{1, 4, 9, 16\}$. We chose the period of the multilayer stack to be 5 times the free-space wavelength at the operating frequency ($\omega/c=2\pi; \Lambda=5$). The increase of the complexity of the electromagnetic structure leads to an increase in the matrix size $\hat A$, making the solution of the eigenvalue problem both computationally and memory-intensive. Note that the majority of eigenvalue solvers rely on iterative algorithms and are therefore not easily deployable in GPU environments. However, a neural network can, in principle, learn to predict the eigenvalues and eigenvectors of the matrix $\hat A$. To demonstrate the scalability of our approach we generate $N=2000$ realizations of the layered structure. For each example, we also generate the associated $\hat{A}$ of size $401 \times 401$ complex values. The combination of the challenging scale of this eigen-decompostion and the scarcity of training data makes this problem interesting from scalability and generalizaility perspective. To demonstrate extrapolation ability, we take a training size $|\mathcal{D}_{Tr}|=370$ realizations that has a refractive index of only 1 in its first layer. On the other hand, we take a test set of size $|\mathcal{D}_{U}|=1630$ with the first layer's refractive index unconstrained (i.e. any value from the set \{1,2,3,4\}). 

\paragraph{Baseline Methods:}
Since there does not exist any related work in PGNN that has been explored for our target applications, we construct analogue versions of \pinn{} \cite{raissi2019physics} and \pgnn{} \cite{karpatne2017physics} adapted to our problem using their major features. We describe these baselines along with others in the following:  
\begin{enumerate}
    \setlength\itemsep{0.0em}
    \item {\nn{} (or NN):} This refers to the ``black-box'' ANN model trained just using Train-Loss without any PG loss terms. 
    \item {\ncmodel:} The analogue version of PGNN \cite{karpatne2017physics} for our problem where the hyper-parameters corresponding to \eloss{} and \nsloss{} are set to a constant value.
    \item {\pinn:} The analogue version of PINN \cite{raissi2019physics} for our problem that performs label-free learning only using PG loss terms with constant weighting coefficients. Note that the PG loss terms are not defined as PDEs in our problem.
    \item {\vpgnn{}:} Multi-task Learning (MTL) variant of PGNN where PG loss terms are optimized alternatively \cite{AlternativeOptimization} by randomly selecting one from all the loss terms for each mini-batch in every epoch.
    \item { \textit{GradNorm} \cite{chen2018gradnorm}: A recent MTL framework developed by Chen et al. \cite{chen2018gradnorm} to adaptively balance loss terms using normalization techniques.}
    \item { Sener et al. \cite{sener2019multitask}: A  state-of-the-art MTL framework developed by Sener et al. \cite{sener2019multitask} that leverages the idea of multi-objective optimization.}
\end{enumerate}

We also consider the following ablation models:
\begin{enumerate}
    \setlength\itemsep{0.0em}
    \item {\nexmodel:} This is an ablation model where the PG loss terms are only trained over the training set, $\mathcal{D}_{Tr}$. Comparing our results with this model will help in evaluating the importance of using unlabeled samples $\mathcal{D}_{U}$ in the computation of PG loss.
    \item {\nemodel:} This is another ablation model where we only consider \nsloss{} in the learning objective, while discarding \eloss{}.
    \item {\lfmodel:} This ablation model drops Train-MSE from the learning objective and hence performs label-free (LF) learning only using PG loss terms.
\end{enumerate}

\paragraph{Evaluation Metrics:}
{
We use four different types of evaluation metrics to assess the predictive accuracy of $\hat{\eigenvec}$ w.r.t. ground-truth $\eigenvec$ in our two target application domains, as described in the following. All metrics are averaged across all test samples. 
\begin{enumerate}
    \setlength\itemsep{0.0em}
    \item {Test-MSE}: In the quantum mechanics application, we considered the mean-squared-error of   predictions on every test sample (i.e., $||\hat{\eigenvec} - \eigenvec||^2$) as an evaluation metric. Since Test-MSE measures the Euclidean distance between $\hat{\eigenvec}$ and ${\eigenvec}$ on test samples, it provides an absolute sense of the scale of errors in the predictions of comparative methods w.r.t ground-truth.
    \item Cosine Similarity: Another evaluation metric that we used for the quantum mechanics application is the cosine similarity between $\hat{\eigenvec}$ and $\eigenvec$, namely, $\langle \frac{\hat{\eigenvec}}{||\hat{\eigenvec}||} , \frac{\eigenvec}{||\eigenvec||}\rangle$, where $\langle \cdot, \cdot \rangle$ denotes the inner product. There are several reasons why cosine provides a more meaningful measure of predictive performance in the problem of solving eigenvalue equations as compared to Test-MSE. First, since eigenvectors are scalable (i.e. if $\eigenvec{}$ is an eigenvector then $a\eigenvec{}$ is also an eigenvector), we want a metric that is invariant to scaling such as cosine. In particular, we want to measure the alignment between $\hat{\eigenvec}$ and $\eigenvec$ across all components of the vector irrespective of their scales. While cosine is suitable for such an analysis, Test-MSE can get affected by the differences in scales of the components of $\eigenvec$ such that the small ones are ignored while the large ones dominate the score. Second, cosine can be used even in high-dimensional settings of $\eigenvec$ (such as those explored in our experiments) where Euclidean distances may lose meaning due to the curse of dimensionality. Finally, an ideal cosine similarity of 1 provides an interpretable baseline to evaluate the goodness of test predictions. A cosine similarity of 0 means that the prediction is orthogonal to the ground-truth eigenvector.
    \item Overlap Integral: Similar to the cosine similarity of real-valued vectors in the quantum mechanics application, we consider the `overlap integral' of complex-valued vectors in the electromagnetic propagation application to measure the alignment between $\hat{\eigenvec}$ and $\eigenvec$ regardless of their scales. In particular, the overlap integral between $\hat{\eigenvec}$ and $\eigenvec$ is defined as $\langle \frac{\hat{\eigenvec}}{||\hat{\eigenvec}||}, \frac{\eigenvec^*}{||\eigenvec^*||}\rangle$, where $\eigenvec^*$ is the complex conjugate of the ground-truth eigenvector. The value of the overlap integral ranges from zero, which means no overlap (or orthogonal), to 1, which means an exact overlap. An overlap integral of $0.8$ or above is commonly desired in the application domain. The overlap integral is also tied to the physical concept of measuring the accuracy of our magnetic field profile predictions. Therefore, it facilitates testing whether our predicted vectors are valid eigenvectors from a physical standpoint.
    \item Relative Eigenequation Error: While both the cosine and overlap integral measure the closeness of the eigenvector's profile to the ground truth, it is also important to understand whether a prediction that does not match with the ground truth is still a valid eigen-solution or not. For this purpose, we use the relative eigenequation error, which is described as $\frac{||\eigenmat \hat{\eigenvec} - \hat{\eigenval} \hat{\eigenvec}||}{||\eigenmat \hat{\eigenvec}||}$. This unitless quantity measures the error in the eigenvalue equation relative to the quantity on the left-hand side of the equation. A value close to zero is desired, indicating that the prediction is a valid eigenvector. Note that a value of 0 can be obtained even if the prediction is not the desired eigenvector (i.e., with the lowest or highest eigenvalue), in which case the cosine or overlap integral will also be equal to 0 because of  orthogonality with the desired eigenvector.

\end{enumerate}

}


\section{Results and analysis}
\label{sec:results}

\subsection{Quantum Mechanics Application}
\label{sec:quantum}



\begin{table}
\caption{Test-MSE and Cosine Similarity of comparative ANN models on training size $N=1000$ on the quantum mechanics application.
}
\resizebox{0.45\columnwidth}{!}{\begin{tabular}{l|rr}
\thickhline
 Models & MSE ${(\times10^{2})}$ & Cosine Similarity \\ \hline
\proposed{} \textbf{(proposed)} & \bm{$0.35\pm0.12$} & \bm{$99.50\pm0.12$}\% \\ \hline

\nn{} & $1.06\pm0.16$ & $95.32\pm0.58$\% \\ 
\pinn{} & $6.27\pm6.94$ & $87.37\pm12.87$\% \\ 
\pgnn{} & $0.91\pm1.90$ & $97.97\pm4.89$\% \\
\vpgnn{} & $6.33\pm2.69$ & $84.26\pm6.33$\% \\ 
\textit{GradNorm} \cite{chen2018gradnorm} & $6.01\pm1.05$ & $89.53\pm4.21$\% \\
Sener et al. \cite{sener2019multitask} & $1.15\pm0.18$ & $94.98\pm0.73$\% \\ \hline
\nexmodel{} & $1.82\pm0.36$ & $93.61\pm0.91$\% \\ 
\nemodel{} & $10.97\pm0.71$ & $76.27\pm0.80$\% \\ 
\lfmodel{} & $9.97\pm4.42$ & $63.97\pm16.20$\% \\ \thickhline
\end{tabular}}
\captionsetup{width=.45\textwidth, font=small}
\label{tab:compare_baseline}
\end{table}

\begin{figure}
    \centering
    \includegraphics[width=0.45\textwidth]{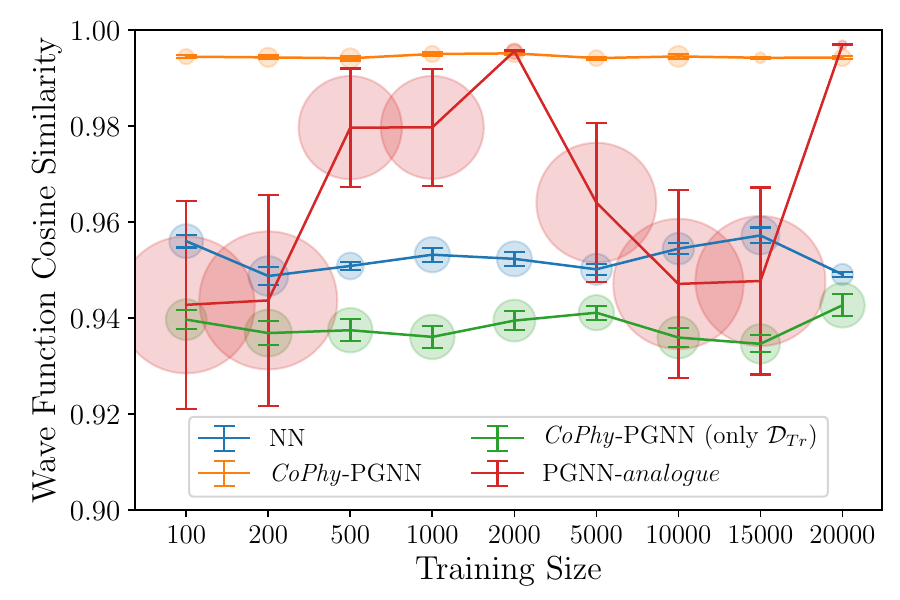}
    \captionsetup{width=.45\textwidth, font=small} 
    \captionof{figure}{Cosine similarity across training sizes. 
    }
    \label{fig:different_train_size}
\end{figure}

Table \ref{tab:compare_baseline} provides a summary of the comparison of \proposed{} with baseline methods on the quantum mechanics application. We can see that our proposed model shows significantly better performance in terms of both Test-MSE and Cosine Similarity. In fact, the cosine similarity of our proposed model is almost 1, indicating close to perfect fit with test labels. (Note that even a small drop in cosine similarity can lead to cascading errors in the estimation of other physical properties derived from the ground-state wave-function.) 
{
On the other hand, the two state-of-the-art baselines in multi-task learning  \cite{chen2018gradnorm, sener2019multitask} produce worse results than  the black-box NN. This is because both these baselines attempt to balance the contribution of every loss term with equal importance during all stages of ANN training, e.g., by balancing the norms of the gradients in the GradNorm method \cite{chen2018gradnorm}, or by performing gradient descent in a direction that favors all of the objectives \cite{sener2019multitask}. While these approaches are useful when the multi-task loss terms are in agreement with each other, they can easily get distracted in the presence of competing gradients of PG loss terms as encountered in our scientific applications. We observed that both these methods struggled to find the lowest eigenvalue solution because it is easy to get stuck at a poor local minima of \nsloss{}. In contrast to balancing PG loss terms with equal importance, we exploit the differences in the characteristics of \nsloss{} and \eloss{} in \proposed{} to favor one over another at different stages of ANN training.


}

An interesting observation from Table \ref{tab:compare_baseline} is that  \lfmodel{} actually performs even worse than black-box NN. This shows that solely relying on PG loss without considering Train-MSE is fraught with challenges in arriving at a generalizable solution. Indeed, using a small number of labeled examples to compute Train-MSE provides a significant nudge to ANN learning to arrive at more accurate solutions. Another interesting observation is that \nexmodel{} again performs even worse than \nn{}. This demonstrates that it is important to use unlabeled samples in $\mathcal{D}_{U}$, which are representative of the test set, to compute the PG loss. 
Furthermore, notice that \nemodel{} actually performs worst across all models, possibly due to the highly non-convex nature of \nsloss{} function that can easily lead to local minima when used without \eloss{}. This sheds light on another important aspect of PGNN that is often over-looked, which is that it does not suffice to simply add a PG-Loss term in the objective function in order to achieve generalizable solutions. In fact, an improper use of PG Loss can result in worse performance than a black-box model. 


\subsubsection{Effect of varying training size:}
\label{sec:results1}

Fig. \ref{fig:different_train_size} shows the differences in performance of comparative algorithms as we vary the training size from $100$ to $20000$. We can see that  \ncmodel{}, which does not perform adaptive tuning, shows a high variance in its results across training sizes. This is because without cold starting $\lambda_C$, \nsloss{} can be quite unstable in the early epochs and can guide the gradient descent into one of its many local minima, especially when the gradients of train-MSE are weak due to paucity of training data. On the other hand, \proposed{} performs consistently better than all other baseline methods, with smallest variance in its results across 10 random runs. In fact, our proposed model is able to perform well even over 100 training samples. 

\subsubsection{Studying convergence across epochs:}
\label{sec:results2}

\begin{figure*}
    \centering
    \hspace{-1ex}
    \includegraphics[width=0.33\textwidth]{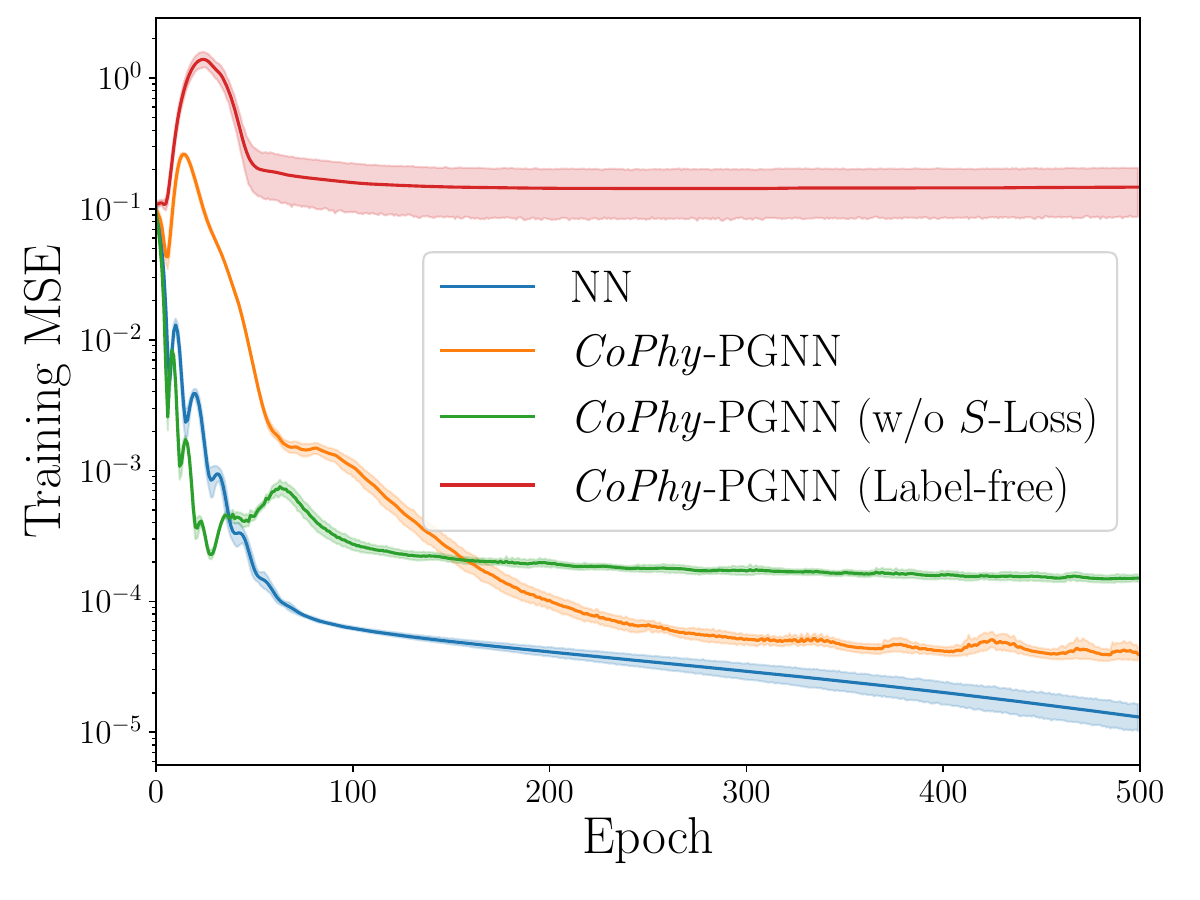}
    \includegraphics[width=0.33\textwidth]{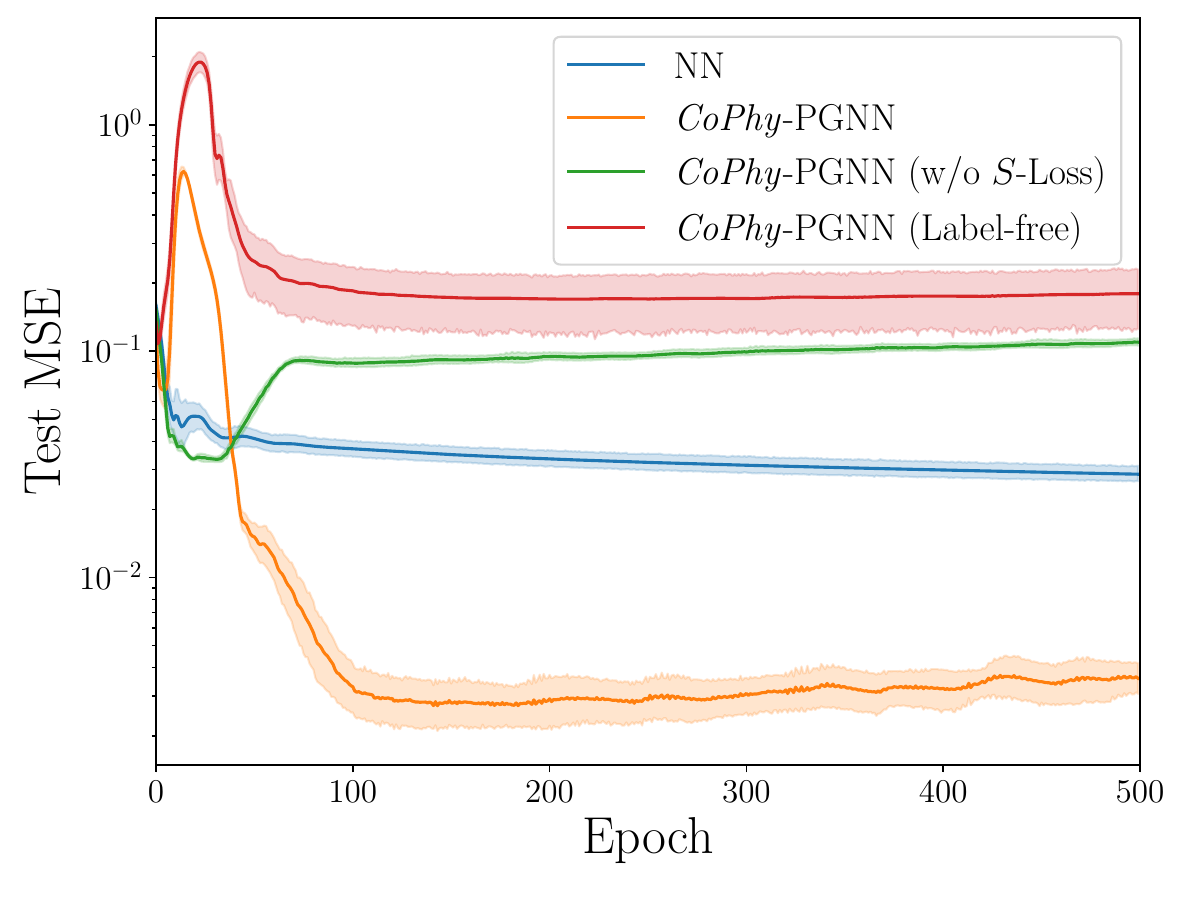}
    \includegraphics[width=0.33\textwidth]{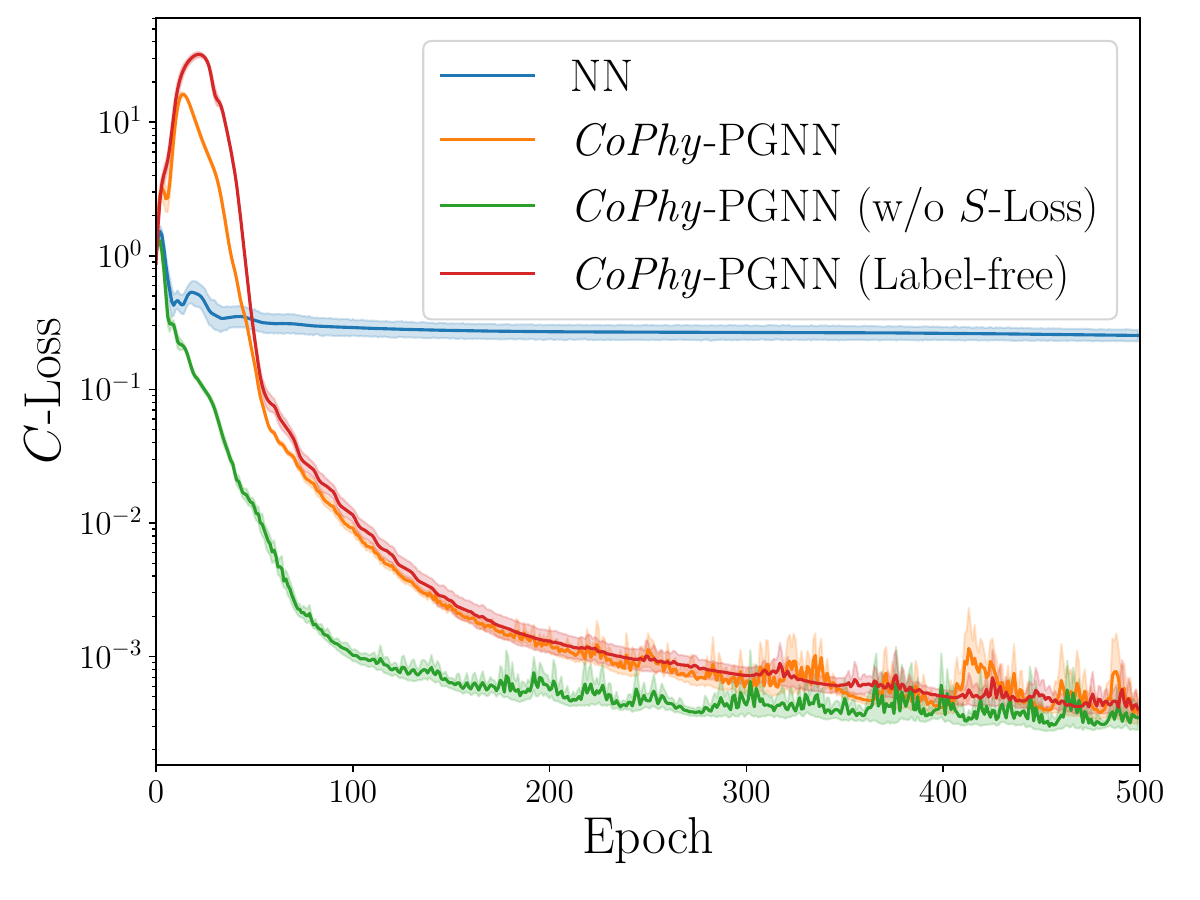}
    \caption{Convergence plots showing Train-MSE, Test-MSE, and C-Loss over epochs. 
    }
    \label{fig:convergence_plot}
\end{figure*}

Figure \ref{fig:convergence_plot} shows the variations in Train-MSE, Test-MSE, and \nsloss{} terms for four comparative models at every epoch of gradient descent. We can see that all models are able to achieve a reasonably low value of Train-MSE at the final solution expect \lfmodel{}, which is expected since it does not consider minimizing Train-MSE in the learning objective. \nn{} actually shows the lowest value of Train-MSE than all other models. However, the quantity that we really care to minimize is not the Train-MSE but the Test-MSE, which is indicative of generalization performance. We can see that while our proposed model, \proposed{} shows slightly higher Train-MSE than \nn{}, it shows drastically smaller Test-MSE at the converged solution, demonstrating the effectiveness of our proposed approach.

\begin{figure}
    \begin{center}
        \includegraphics[width=0.38\textwidth]{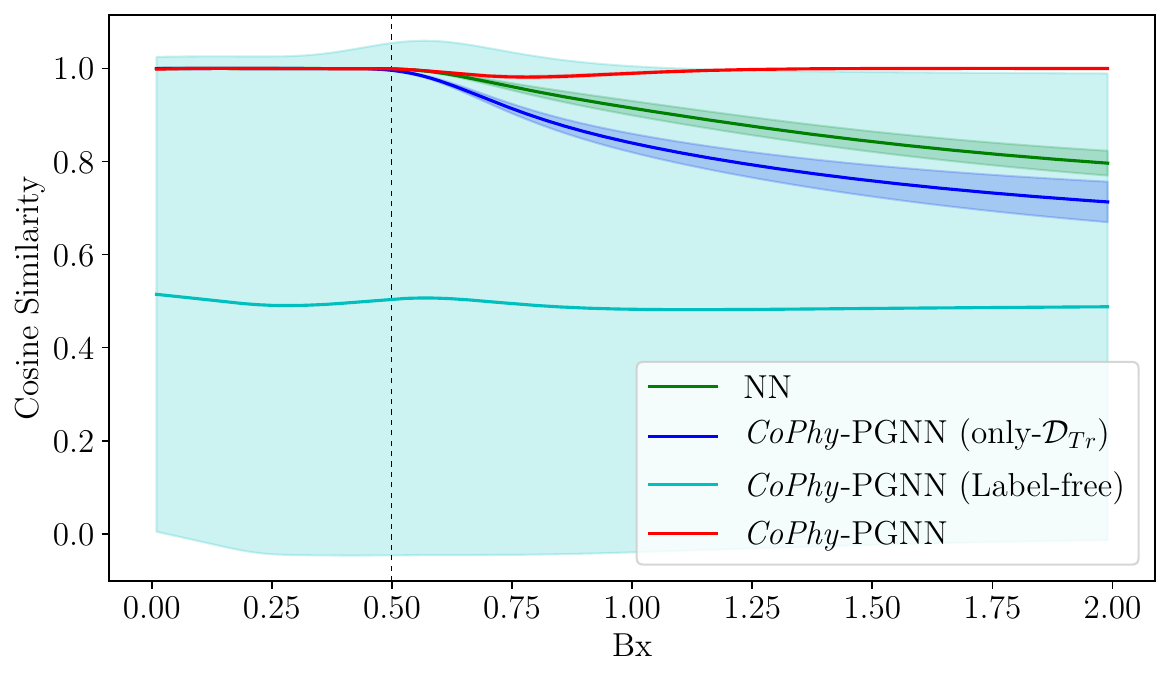}
    \end{center}
    \captionsetup{font=small}
    \caption{{Cosine Similarity on test samples as a function of $B_x$. The dashed line represents the boundary between the interval used for training (left)  and  testing (right).}}
    \label{fig:cosineSimilarity}
\end{figure}


A contrasting feature of the convergence plots of \proposed{} relative to \nn{} is the presence of an initial jump in the Test-MSE values during the first few epochs. This likely arises due to the competing nature of two different loss terms that we are trying to minimize in the beginning epochs: the Train-MSE, that tries to move towards local minima solutions favorable to training data, and \eloss{}, that pushes the gradient descent towards generalizable solutions. Indeed, this initial jump in Test-MSE helps in moving out of local minima solutions, after which the Test-MSE plummets to significantly smaller values. Notice that \lfmodel{} shows a similar jump in Test-MSE in the beginning epochs, because it experiences a similar effect of \eloss{} gradients during the initial stages of ANN learning. However, we can see that its Test-MSE is never able to drop beyond a certain value after the initial jump, as it does not receive the necessary gradients of Train-MSE that helps in converging towards generalizable solutions. 

Another interesting observation is that \nemodel{} does not show any jump in Test-MSE during the beginning epochs in contrast to our proposed model, since it is not affected by \eloss{}. If we further look at \nsloss{} curves, we can see that \nemodel{} achieves lowest values, since it only considers \nsloss{} as the PG loss term to be minimized in the learning objective. However, we know that \nsloss{} is home to a large number of local minima, and for that reason, even though \nemodel{} shows low \nsloss{} values, its test-MSE quickly grows to a large value, indicating its convergence on a local minima. These results demonstrate that a careful trade-off of PG loss terms along with Train-MSE is critical to ensure good generalization performance, such as that of our proposed model.
To better understand the behavior of competing loss terms, we conducted a novel gradient analysis that can be found in Section \ref{sec:gradients}.

\subsubsection{Evaluating generalization power:}
\label{sec:results3}

Instead of computing the average cosine similarity across all test samples, Figure \ref{fig:cosineSimilarity} analyzes the trends in cosine similarity over test samples with different values of $B_x$, for four comparative models. Note that  
none of these models have observed any labeled data during training outside the interval of $B_x \in [0,0.5]$. Hence, by testing for the cosine similarity over test samples with $B_x > 0.5$, we are directly testing for the ability of ANN models to generalize outside the data distributions it has been trained upon. Evidently, all label-aware models perform well on the interval of $B_x\in [0,0.5]$. However, except for \proposed{}, all baseline models degrade significantly outside that interval, proving their lack of generalizability. Moreover, the label-free, \lfmodel{}, model is highly erratic, and performs poorly across the board.


\subsubsection{Analysis of loss landscapes:}
\label{section:lossSection}
\begin{figure*}
    \centering
    \includegraphics[width=0.9\textwidth]{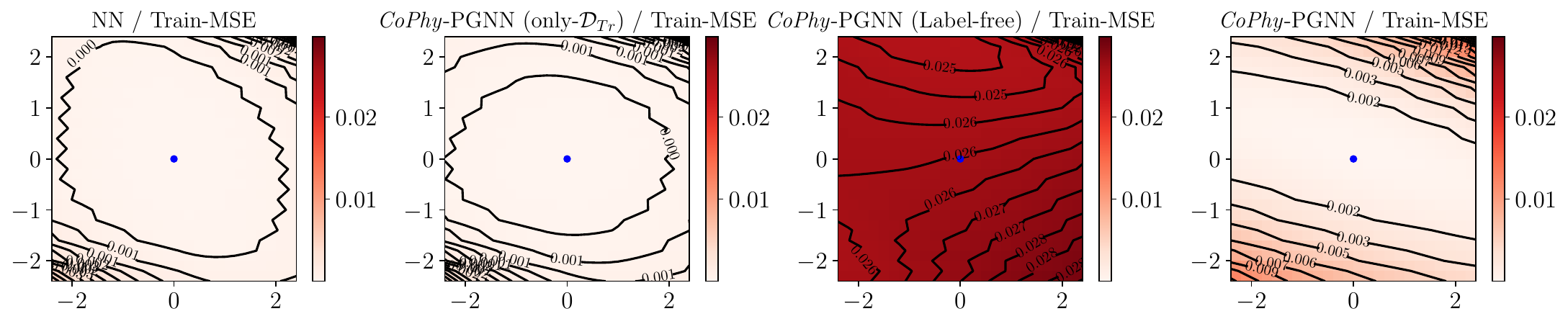}
    \includegraphics[width=0.9\textwidth]{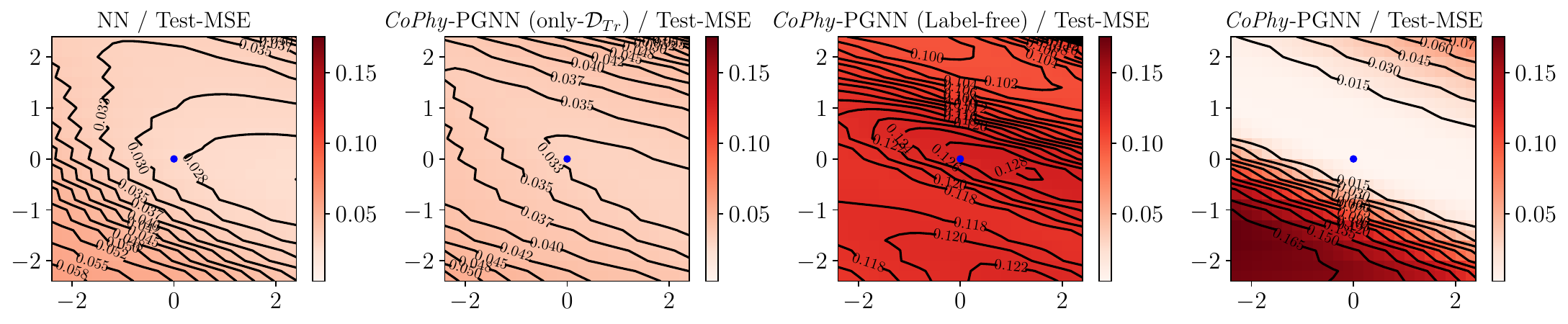}
    \includegraphics[width=0.9\textwidth]{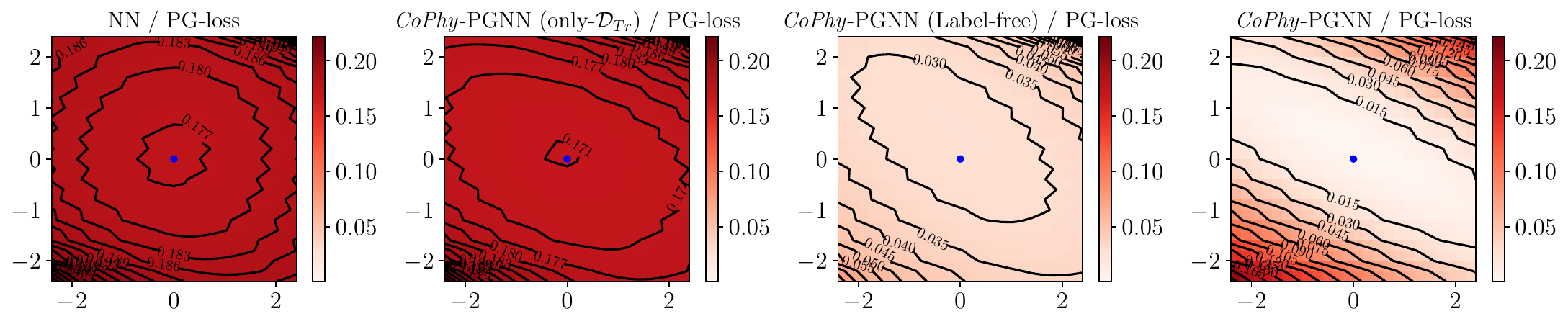}
\captionsetup{font=small}
\caption{A comprehensive comparison between \proposed{} and different baselines. The 1st and 2nd columns show that without using unlabeled data, the model does not generalize well. On the other hand, the 3rd column shows that without labeled data, the model fails to reach a good minimum. Only the last column, our proposed model, shows a good fit across both labeled and unlabeled data. The best performing model is also the model that best optimizes the PG loss.}
\label{fig:CMT_surfaces}
\end{figure*}

To truly understand the effect of adding PG loss to ANN’s generalization performance, here we visualize the landscape of different loss functions w.r.t. ANN model parameters. 
In particular, we use the code in \cite{loss-landscapes} to plot a 2D view of the landscape of different loss functions, namely Train-MSE, Test-MSE, and PG-Loss (sum of \nsloss{} and \eloss{}), in the neighborhood of a model solution, as shown in Figure \ref{fig:CMT_surfaces}. In each of the sub-figures of this plot, the model's parameters are treated with filter normalization as described in \cite{NIPS2018_7875}, and hence, the coordinate values of the axes are unit-less. Also, the model solutions are represented by blue dots. As can be seen, all label-aware models have found a minimum in Train-MSE landscape. However, when the test-MSE loss surface is plotted, it is clear that while the \proposed{} model is still at a minimum, the other baseline models are not. This is a strong indication that using the PG loss with unlabeled data can lead to better extrapolation; it allows the model to generalize beyond in-distribution data. 
We can see that without using labels, \lfmodel{} fails to reach a good minimum of Test-MSE, even though it arrives at a minimum of PG Loss.

\subsubsection{Analysis of gradients}
\label{sec:gradients}

\begin{figure}[ht]
    \centering
    \includegraphics[width=0.45\textwidth]{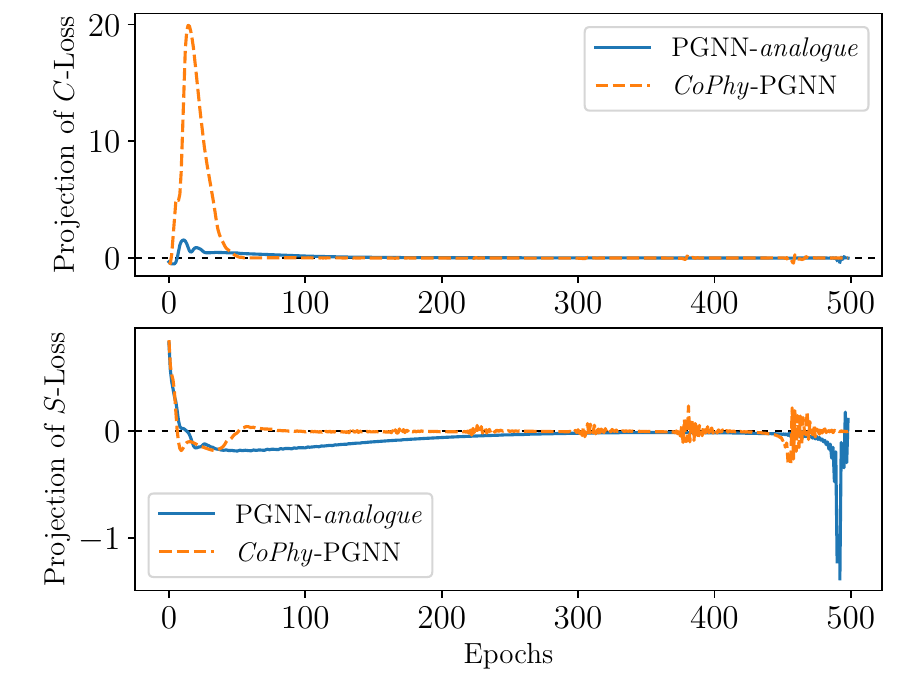}
    \caption{The projection of gradient of each term on the optimal direction. The optimal direction of a certain iteration points from current state to the optimal state.}
    \label{fig:gradient_analysis}
\end{figure}

The complicated interactions between competing loss terms motivate us to further investigate the different role each loss term plays in the aggregated loss function. The sharp bulge in both \mse{} and \nsloss{} in the first few epochs, as shown in Fig \ref{fig:gradient_analysis},
shows that the optimization process is not quite smooth. Our speculation is that \eloss{} and \nsloss{} are competing loss terms in the combined optimization problem. 
To monitor the contribution of every loss term in the learning process, we need to measure if the gradients of a loss term points towards the optimal direction of descent to a generalizable model.
One way to achieve this is to compute the component of the gradient of a loss term in the optimal direction of descent (leading to a generalizable model). Suppose the desired (or optimal) direction is $\bm{d}^*$ and the gradient of a loss term $L$ is $\nabla{L}$. We can then compute the projection of $\nabla{L}$ along the direction of  $\bm{d}^*$ at the $k^\text{th}$ epoch as: 
\begin{equation}
    p_L^{(k)} = \frac{\langle \nabla{L}^{(k)}, \bm{d}^{*(k)}\rangle}{\|\bm{d}^{*(k)}\|}.
\end{equation}

A higher projection value indicates a larger step toward the optimal direction at the $k^\text{th}$ epoch, $\bm{d}^{*(k)}$, which is defined as:
\begin{equation}
    \bm{d}^{*(k)} = \bm{\theta}^{(k)} - \bm{\theta}^*,
    \label{eq:opt_direction}
\end{equation}

where $\bm{\theta}^{(k)}$ is the model parameters $\bm{\theta}$ (i.e., weights and biases of the neural network) at the $k^\text{th}$ epoch and $\bm{\theta}^*$ is the optimal state of the model that is known to be generalizable. Note that finding an exact solution for $\bm{\theta}^*$ that is the global optima of the loss function is practically infeasible for deep neural networks \cite{LossSurfaceLocalMinima}. Hence, in our experiments, we consider the final model arrived on convergence of the training process as a reasonable approximation of $\bm{\theta}^*$.
For methods such as \ncmodel{} and \proposed{}, the final models at convergence performed significantly well and showed a cosine similarity of $\geq 99.5\%$ with ground-truth. This is very close to a model trained directly on the test set that only reaches 99.8\%. This gives some confidence that the final models at convergence are good approximations to $\bm{\theta}^*$. To compute the inner products between $\nabla{L}$ and $\bm{d}^*$, we used a flattened representation of the model parameters by concatenating the weights and biases across the layers.


We analyze the role of \eloss{} and \nsloss{} in the training process of the two methods: \proposed{} and \pgnn{}. Both these methods were run with the same initialization of model parameters. 
Training size is 2000 and the rest of settings are the same as in Section 5.

Figure \ref{fig:gradient_analysis} shows that in the early epochs, the \eloss{} has positive projection values, which means that it is helping the method to move towards the optimal state. On the other hand, the projection of \nsloss{} starts with a negative value, indicating that the gradient of the \nsloss{} term is counterproductive at the beginning. Hence, \eloss{} helps in moving out of the neighborhood of the local minima of \nsloss{} towards a generalizable solution. 
However, the projection of \nsloss{} does not remain negative (and thus counterproductive) across all epochs. In fact, \nsloss{} makes a significant contribution by having a large positive projection value after around 50 epochs. This shows that as long as we manage to escape from the initial trap caused by the local minima of the \nsloss{}, then it can turn to guide the model towards desired direction $\bm{d}^{*}$. By initially setting $\lambda_C$ close to zero, it allows the \eloss{} to dominate in the initial epochs and move out of the local minima. Later, we let \nsloss{} recover to a reasonable value and it will start to play its role. These findings align quite well with the \emph{cold-start} and \emph{annealing} idea proposed in this work and show that it works best when the two loss terms are combined together using adaptive weighting coefficients. 



Note that for this analysis method to produce valid findings, we need to ensure that the loss terms are not pointing towards the direction of an equally good $\bm{\theta}^*$ that can be arrived at from the same initialization. To ensure this, we investigated how similar the trained models (optimal states) are when started from the same initialization for the two methods. The parameters of the \ncmodel{} and \proposed{} showed an average cosine similarity of 98.6\%, and in many cases reached 99\%. This gave us more confidence to believe that our approximations to the optimal model were sufficient. 

\subsection{Electromagnetic Propagation Application}
\label{sec:propagation}
{
While the relatively smaller size of input matrices in the quantum mechanics problem helped in performing detailed empirical comparisons of \proposed{} with respect to baselines, our experiments on the electromagnetic propagation problem are geared toward demonstrating the scalability of our approach to larger systems while ensuring generalizability. In particular, note that the size of every complex-valued matrix in this application is $401 \times 401$, making the scale of this problem about 2500 times larger than that of the quantum mechanics problem. Also, the the labeled set is restricted to 370 samples with permittivity $\epsilon=1$ on the first layer, making the learning problem significantly more challenging for any neural network approach. Given the computational demands of this problem, we did not perform an exhaustive grid search of the hyper-parameters for all comparative methods. Instead, we were able to find a reasonable setting of hyperparameters of \proposed{} by making adjustments to their default values from the quantum mechanics application and observing the optimization process on the training data. Also, in order to account for the presence of complex values in this application, we made a minor modification to the label loss by using the overlap integral instead of MSE on the labeled training set. Details of the hyper-parameter selection, the modified label loss, and other architectural specifications are provided in Appendix \ref{sec:em-appendix}.






\subsubsection{Evaluating Generalization Power:}
To assess the extrapolation power of \proposed{} on out-of-distribution test samples, we plot the relative eigenequation error for comparative methods over test samples with different values of permittivity $\epsilon$ of the first layer in Figure \ref{fig:electromagnetic}. Note that the labeled set used for training only included samples with $\epsilon = 1$ (the leftmost point in the curves of Figure \ref{fig:electromagnetic}), while the test set included samples with $\epsilon \in \{1,4,9,16\}$. In this extrapolation experiment, we compared the performance of \proposed{} w.r.t. two baseline methods: NN and \ncmodel{}. We particularly chose the \ncmodel{} because it was the top-performing baseline method in the quantum mechanics application. We can see that both these baselines show lowest relative errors at $\epsilon = 1$, while there is a significant jump in their relative errors as we go beyond $\epsilon > 1$. This behavior is expected since all methods were trained on labeled samples with $\epsilon$ only equal to 1. However, we can notice that in contrast to the baseline methods, \proposed{} shows the lowest relative error across all values of $\epsilon$ and does not show any apparent increase in relative error as we go beyond $\epsilon > 1$. This again demonstrates the ability of our proposed framework to generalize on unseen test distributions by making effective use of PG loss terms during ANN training.


\begin{figure}
\centering
    \includegraphics[width=0.40\textwidth]{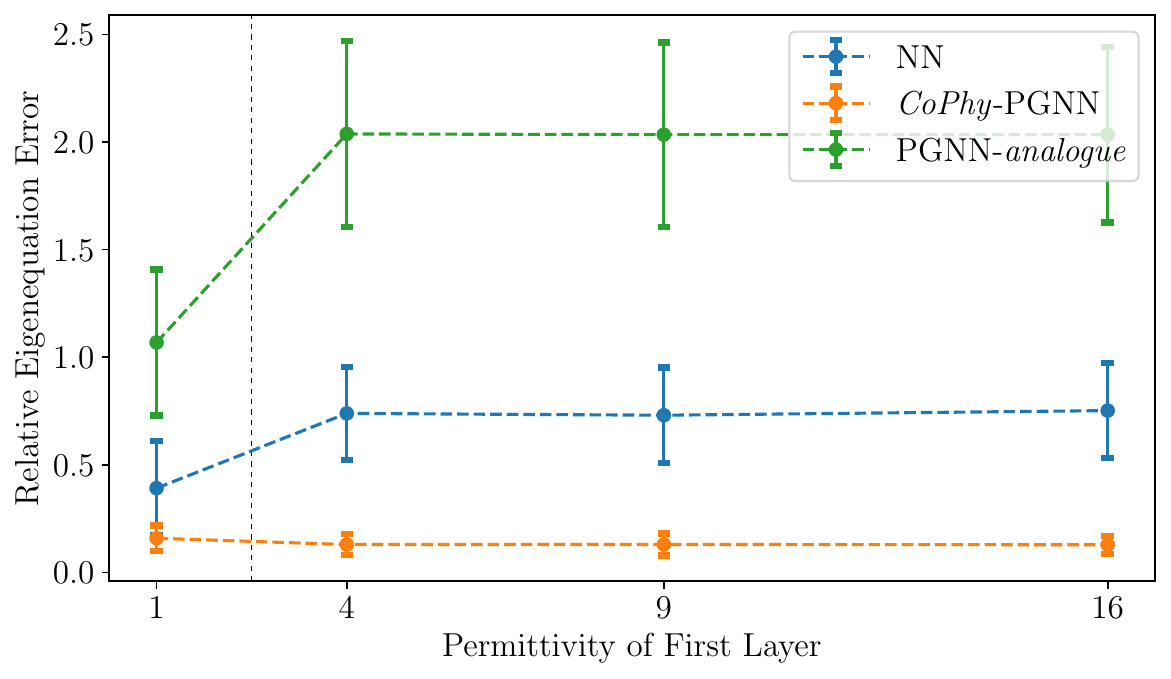}
    \captionsetup{font=small}
    \caption{The relative eigenequation error of \proposed{} compared to \nn{} and \ncmodel{} for the electromagnetic propagation application as a function of the permittivity of the first layer, $\epsilon$. The dashed black line represents the boundary between the interval used for training (left) and  testing (right).}
\label{fig:electromagnetic}
\end{figure}

\subsubsection{Analyzing Predictive Accuracy vs. Runtime Trade-off:}
One of the major advantages of using an ANN approach for solving eigenvalue problems is that while training the ANN model can be time-consuming, it can provide significant improvements in runtime during testing as compared to numerical methods. This is because ANN models, once trained, are very fast at making forward inferences using simple matrix operations that are amenable to modern GPU architectures, in contrast to numerical solvers that perform computationally expensive operations on every test sample. In particular, ANN models are especially useful in scientific applications where we need to make predictions on large batches of input samples during testing. However, black-box ANN models, although fast, can produce inaccurate solutions on unseen test samples, especially when the test samples are generated from distributions different than those encountered during training. We analyze the ability of \proposed{} to balance the twin objectives of improving predictive accuracy compared to baseline ANN models and increasing runtime speed compared to numerical solvers in the application of electromagnetic propagation.



\begin{table}
\caption{Comparison of testing speed and predictive accuracy between \proposed{} and other ANN methods and numerical  eigen-solvers. Note that Matlab's results are the most accurate as the dataset itself was generated using Matlab. Also note that the runtime of all ANN based solvers is identical but their predictive accuracy values are different.}
\resizebox{0.75\columnwidth}{!}{
{
\begin{tabular}{llll}
\thickhline
Solver & average runtime (seconds) & relative eigenequation error & overlap integral \\ 
\hline
\proposed{} & {$4.3 \times 10^{-4}$}                 & $1.309 \times 10^{-1}$  & $0.645$   \\
\ncmodel{} &    {$4.3 \times 10^{-4}$}                                                    & $1.980 \times 10^{-0}$  & $0.616$   \\
\nn{} &                               {$4.3 \times 10^{-4}$}                              & $7.205 \times 10^{-1}$  & $0.579$   \\
\hline
\texttt{numpy.linalg.eig}  \cite{numpy}   & $3.332 \times 10^{1}$                 & $1.466 \times 10^{-7} $ & $0.999$ \\
\texttt{Matlab} \cite{matlab}  & $1.482 \times 10^{-1}$              & $2.352 \times 10^{-10}$  &  $1.000$  \\
\texttt{scipy.linalg.eig}  \cite{scipy}  & $3.225\times 10^{-1} $                & $2.620  \times 10^{-5}$ & $0.982$ \\
\thickhline
\end{tabular}}}
\captionsetup{width=.5\textwidth, font=small}
\label{table:Speed}
\end{table}


Table \ref{table:Speed} compares the performance of different methods (both ANN-based and numerical solvers) in terms of average runtime of testing, the overlap integral of predictions, and the relative eigenequation error of predictions, averaged over the entire test set. Note that the runtime of all ANN based approaches (namely, \proposed{}, \ncmodel{}, and \nn{}) is identical because of the exact same nature of computations performed during testing. However, they show different predictive accuracy values. We can see that \proposed{} shows the lowest relative eigenequation error and the highest overlap integral among all ANN based methods. On the other hand, the runtime of \proposed{} is about three three orders of magnitude smaller than any numerical eigen-solver. This demonstrates that our proposed framework is able to strike the right balance of showing the best predictive accuracy among ANN methods while enjoying significant speed-ups in runtime compared to numerical methods. While these results of \proposed{} mark an important milestone on the path to reducing the accuracy gap between ANN models and numerical solvers, they are only the first step towards realizing the full potential of physics-guided neural networks for solving eigenvalue problems. In particular, there is a lot of scope for improving the predictive accuracy of \proposed{} w.r.t. numerical methods, e.g., by performing an informed search of its hyper-parameters or by increasing the size and representation power of the labeled training set.


Note that the drop in the  relative eigenequation error of \proposed{} w.r.t. ANN methods is much greater than the jump in its overlap integral metric. To further investigate this trend, we decided to plot the histogram of overlap integrals for all three ANN methods on the test set in Figure \ref{fig:histograms}. We can see that most of \proposed{}'s predictions are either mostly aligned perfectly well with the ground-truth, with an integral overlap value of 1, or barely has any overlap with it, exhibiting an integral overlap of 0. This, when combined with its exceptionally low relative eigenequation error shown in Figure \ref{fig:electromagnetic}, indicates that \proposed{} surpasses the other ANN baselines in the ability to solve eigenequations in general, but sometimes fails at obtaining the desired (i.e., largest) eigenvector in favor of another valid one. Note that since all eigenvectors are orthogonal to each other, predicting a valid eigenvector that is not equal to the ground-truth would result in an overlap integral of 0  even though its relative eigenequation error would be 0. This is in contrast to other ANN baselines that have significantly fewer examples with almost perfect profile alignment with the ground-truth.


\begin{figure}
    \centering
    \includegraphics[width=0.40\textwidth]{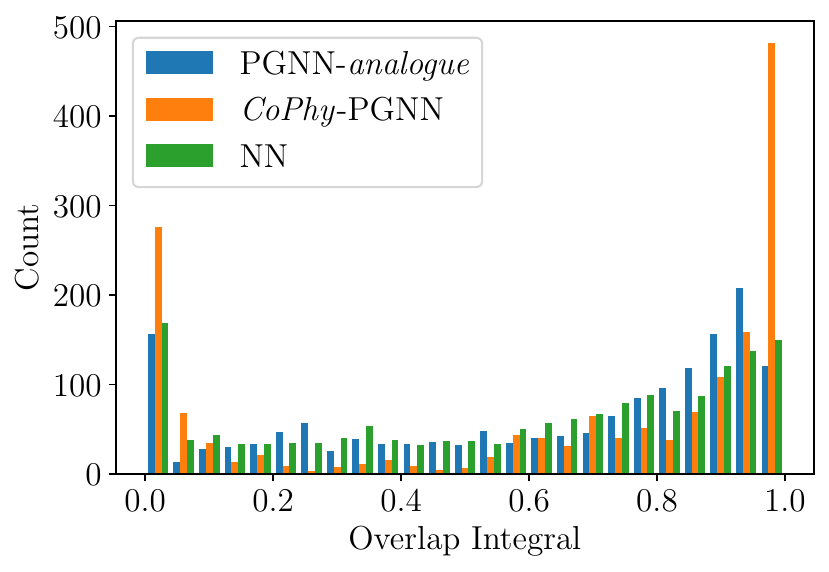}
    \caption{Histogram showing the distribution of the overlap integral of ANN methods over the test set. Here, unlike \nn{} and \ncmodel{}, we can see that most examples return either an almost perfect overlap or barely any overlap for \proposed{}.} 
    \label{fig:histograms}
\end{figure}

\subsubsection{Analyzing the Effect of Batch Size on ANN Scalability:}

The biggest advantage of neural networks such as \proposed{} is their runtime scalability when tested over large batches of samples.
The results shown in Table \ref{table:Speed} demonstrate that ANNs are more computationally efficient than iterative numerical solvers even when they are tested over samples one at a time (i.e., without creating batches of test samples). However, this runtime speed gap would be much larger when ANN are tested in a batch-wise fashion, as investigated in the following.

\begin{figure}
    \centering
    \includegraphics[width=0.40\textwidth]{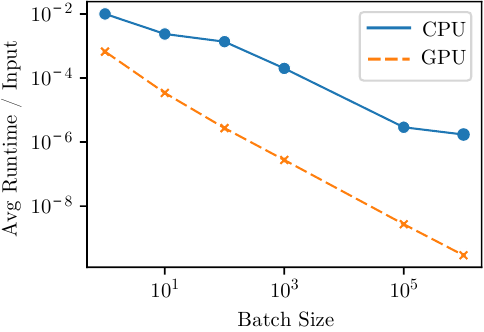}
    \caption{Average runtime per input for \proposed{} w.r.t. different batch sizes (in log-scale).} 
    \label{fig:runtim_scability}
\end{figure}

Figure \ref{fig:runtim_scability} shows the runtime scalability of \proposed{} when the network takes batches of different sizes as inputs during testing. As expected, the average runtime per input decreases dramatically as larger batches are used. In our experiments on a TITAN RTX graphic card, using a batch size of 1M, \proposed{} showed radically better throughput of more than \textbf{300M} inputs per second. On the other hand, the speed of GPU based iterative solvers are far behind ANN methods. This is because while GPUs are known to be able to handle matrix/tensor operations, they are less efficient at handling iterative methods compared to CPUs. Thus, unlike ANNs, iterative algorithms cannot take full advantage of GPUs' parallelization capabilities. This provides sufficient grounds for further investigation in the development of ANN methods for solving eigenvalue problems as explored by \proposed{}.
}
\section{Conclusions and future work}
\label{sec:conclusion}

This paper proposed novel strategies to address the problem of competing physics-guided (PG) loss functions in PGNN. For the general problem of solving eigenvalue equations, we designed a PGNN model named \proposed{} and demonstrated its efficacy in two target applications in quantum mechanics and electromagnetic propagation. 
From our results, we found that: 1) PG loss helps to extrapolate to unseen testing scenarios and imparts better generalizablity to the results; and 2) Using labeled data along with PG loss results in more stable PGNN models. Moreover, we visualized the loss landscape to give a better understanding of how the combination of both labeled data loss and PG loss leads to better generalization performance. 
We have also demonstrated the generalizability of our \proposed{} to multiple application domains with varying types of PG loss functions, as well as its scalability to large systems. 
Future work can focus on reducing the training time of our model so as to perform extensive hyper-parameter tuning to reach a better global minima. Finally, while this work empirically demonstrated the value of \proposed{} in combating with competing PG loss terms, future work can focus on theoretical analyses of our approach.

\section{Acknowledgments}
This work was supported by the National Science Foundation via grants 2026710 (for M. Elhamod, J. Bu, and A. Karpatne), 2026702 (for C. Singh, M. Redell, and W-C. Lee), and 2026703 (for A. Ghosh and V. Podolskiy).

\bibliographystyle{ACM-Reference-Format}

\bibliography{anuj,chris,Moe, Jie}

\clearpage

\appendix

\title{Appendix}

\section{Relevant Physics Background}
\label{sec:background}

\subsection{Ising Chain and Quantum Mechanics}
Quantum mechanics provides a theoretically rigorous framework to investigate physical properties of quantum materials by solving the Schr\"{o}dinger's equation---the fundamental law in quantum mechanics. The Schr\"{o}dinger's equation is essentially a PDE that can be easily transformed into an eigenvalue problem of the form: $\mathbf{H} {\Psi} = E {\Psi}$, where $\mathbf{H}$ is the Hamiltonian Matrix, $\Psi$ is the wave-function, and $E$ is the energy, a scalar quantity. (Note that many other PDEs in physical sciences, e.g., Maxwell's equations, yield to a similar transformation to an eigenvalue problem.)
All information related to the dynamics of the quantum system is encoded in the eigenvectors of $\hat{H}$, i.e., $\Psi$. Among these eigenvectors, the ground state wave-function, $\Psi_0$, defined as the eigenvector with lowest energy, $E_0$, is a fundamental quantity for understanding the properties of quantum systems. Exploring how $\Psi_0$ evolves with controlling parameters, e.g., magnetic field and bias voltage, is an important subject of study in material science.

A major computational bottleneck in solving for the ground-state wave-function $\Psi_0$ is the diagonalization of the Hamiltonian matrix, $\mathbf{H}$, whose dimension grows exponentially with the size of the system. In order to study the effects of controlling parameters on the physical properties of a quantum system, theorists routinely have to perform diagonalizations on an entire family of Hamiltonian matrices, with the same structure but slightly different parameters. 
\begin{figure}[ht]
\centering
\includegraphics[width=0.45\textwidth]{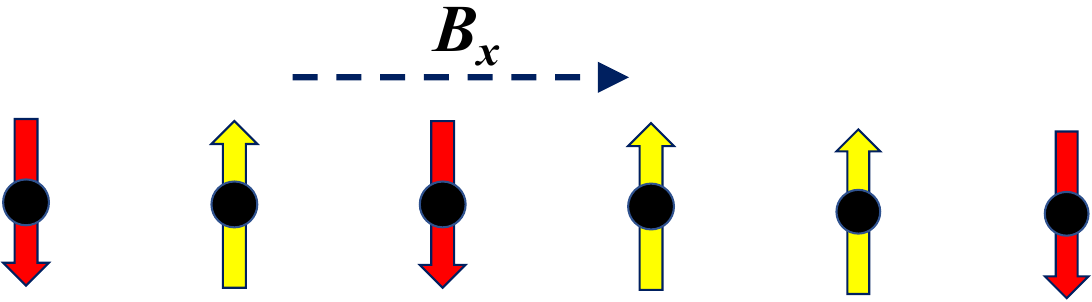}
\caption{{Schematic illustration of the Ising spin chain. Each site is occupied by a spin that can only take two values, either spin up (+1) or spin down (-1). The external magnetic field $B_x$ is applied along the chain direction.}}
\label{fig:spin-chain}
\end{figure}

Here we study a quintessential model of the transverse field: Ising chain model \cite{bonfim2019}, which is
a uni-dimensional spin chain model under the influence of a transverse
magnetic field $B_x$, as shown in Fig. \ref{fig:spin-chain}. 
Spin is the intrinsic angular momentum possessed by elementary particles including electrons, protons, and neutrons. The Ising spin chain model describes a system in which multiple spins are located along a chain and they interact only with their neighbors. By adding an external magnetic field ($B_x$), the ground state wave-function could change dramatically. This model and its derivatives have been used to study a number of novel quantum materials(\cite{tmo,qsl}) and can also be used for quantum computing\cite{qubit}, since the qubit, the basic unit of quantum computing, can also be represented as a spin.
However, the challenge in finding the ground-state wave-function of this model is that the dimension $d$ of the Hamiltonian $\mathbf{H}$ grows exponentially as $d = 2^n$, where $n$ equals the number of spins. We aim to develop PGNN approaches that can learn the predictive mapping from the space of Hamiltonians, $\mathbf{H}$, to ground-state wave-functions, $\Psi$, using the physics of the Schr\"{o}dinger's equation along with labels produced by diagonalization solvers on training set.

\subsection{Electromagnetic Propagation and Maxwell Equations}
Mathematically, the propagation of the electromagnetic waves in periodically stratified layer stacks can be described by Maxwell equations: 
\begin{eqnarray}
\vec\nabla\cross\vec H=\frac{1}{c}\epsilon\frac{\partial \vec E}{\partial t}
\\
\vec\nabla\cross\vec E=-\frac{1}{c}\frac{\partial \vec H}{\partial t},
\end{eqnarray}
where $\vec E$, $\vec H$, $\epsilon$, and $c$ represent electric and magnetic fields, permittivity of layered material, and speed of light in vacuum (in Gaussian units) respectively, and $t$ stands for time. When permittivity of layered array is periodic function of $x$: $\epsilon(x+\Lambda)=\epsilon(x)$, propagation of monochromatic ($\vec E,\vec H\propto e^{-i\omega t})$ transverse-magnetic (TM) polarized waves ($\vec H \|\hat y$) can be reduced to the eigenvalue problem: 
\begin{eqnarray}
\hat A \vec h_m =k^m_z \vec h_m  
\label{eq:electromagnetic}
\end{eqnarray}
where $k_z$ represents the propagation constant of the electromagnetic modes along the layers ($\vec E,\vec H\propto e^{i k_z z}$), $\vec h_m$ represent coefficients of the Fourier transform of the spatial profile of $\vec H(x)$, and the elements of the matrix $\hat A$ is related to operating frequency $\omega$, structure of the electromagnetic waves in the direction normal to the layers (parameterized by quasi-wavenumber $k_x$), and spatial profile of permittivity $\epsilon(x)$ \cite{vpRCWA}. This technique, known as Rigorous Coupled Wave Analysis, has been extended to electromagnetic structures with two dimensional periodicity \cite{vpRCWAnormal} as well as to aperiodic structures \cite{vpRCWAnonperiodic}. 


\section{Additional Experimental Details}
\label{sec:hyperparam}

\subsection{Quantum Mechanics Application}
\subsubsection{Hyper-parameter Search}
To exploit the best potential of the models, we conducted hyper-parameter search prior to many of our experiments\footnote{All the code and data used in this work is available at \url{https://github.com/jayroxis/Cophy-PGNN}. A complete set of code, data, pretrained models, and stored variables can be found at: \url{https://osf.io/ps3wx/?view\_only=9681ddd5c43e48ed91af0db019bf285a} (cophy-pgnn.tar.gz).} on training set of size $N=20000$, by doing random sampling in a fixed range for every hyper-parameter value. We chose the average of the top-5 hyper-parameter settings that showed the lowest error on the validation set, which was 2000 instances sampled from the training set. For the proposed \proposed{} model, this resulted in the following set of hyper-parameters: $\{\lambda_{S0}=2.3, \alpha_S=0.14, \lambda_{C0}=0.85, \alpha_C=0.17, T_a=51\}$, and we chose $T=50$ fixed for all models. The same hyper-parameter values were used across all values of $N$ in our experiments to show the robustness of these values. 

We searched for the best model architecture using simple multi-layer neural networks that does not show significant overfitting or underfitting, then we fixed the architecture for all the models in our work. The models comprise of four fully-connected layers with $tanh$ activation and an linear output layer. The widths of all the hidden states are 100. All the experiments used \textit{Adamax} \cite{kingma2014method} optimizer and set maximum training epochs to 500 that most of the models will converge before that limit. 

Since different models may use different loss terms, the numbers of hyper-parameters to search are different, and some of them were not being searched. We use random search \cite{rand_hyperparam} and run around 300 to 500 runs per model to keep a balance between search quality and the time spent. The hyper-parameters we searched include: 
\begin{enumerate}
    \item For \eloss{}: $\lambda_{C0} \sim \mathcal{U}(0, 5)$, $\alpha_E \sim \mathcal{U}(0, 0.5)$.
    \item For \nsloss{}: $\lambda_{C0} \sim \mathcal{U}(0, 2)$, $\alpha_C \sim \mathcal{N}(0, 0.5)$, $T_{\alpha} \sim \mathcal{U}(0, 200)$
\end{enumerate}

\subsubsection{Sigmoid Cold-start and Other Different Modes}
Additionally, to further prove our choice on $sigmoid$ is indeed effective. We compared three other modes with $sigmoid$: \textit{quick-drop} (Eq. \ref{eq:quick-drop}), \textit{quick-start} (Eq. \ref{eq:quick-start}), \textit{inversed-sigmoid} (Eq. \ref{eq:inverse_sigmoid}).

\begin{equation}
    \lambda_C(t) = \lambda_{C0} \times [1 - sigmoid(\alpha_C (t - T_{\alpha}))]
    \label{eq:inverse_sigmoid}
\end{equation}

\begin{equation}
    \lambda_C(t) = \lambda_{C0} \times (1 + \alpha_C)^{\min(0, - t + T_{\alpha})}
    \label{eq:quick-drop}
\end{equation}

\begin{equation}
    \lambda_C(t) = \lambda_{C0} \times [1 - (1 + \alpha_C)^{\min(0, - t + T_{\alpha})}]
    \label{eq:quick-start}
\end{equation}

We replaced the $sigmoid$ cold-start with the three modes in \proposed{} and ran 400 times for each mode to do hyper-parameter search on $\lambda_{C0}$, $\alpha_C$, $T_{\alpha}$. Using the average hyper-parameter values for the top-5 models that have the lowest validation error. The results are:
\begin{enumerate}
    \item \textit{quick-drop}: $\lambda_{C0}=0.836881$, $\alpha_C=0.062851$, $T_{\alpha}=14.0$.
    \item \textit{quick-start}: $\lambda_{C0}=0.936669$, $\alpha_C=0.073074$, $T_{\alpha}=61.2$.
    \item \textit{sigmoid}: $\lambda_{C0}=0.846349$, $\alpha_C=0.020170$, $T_{\alpha}=51.0$.
    \item \textit{inversed-sigmoid}: $\lambda_{C0}=0.939779$, $\alpha_C=0.171778$, $T_{\alpha}=59.2$.
\end{enumerate}

\begin{figure}[t]
    \centering
    \includegraphics[width=0.45\textwidth]{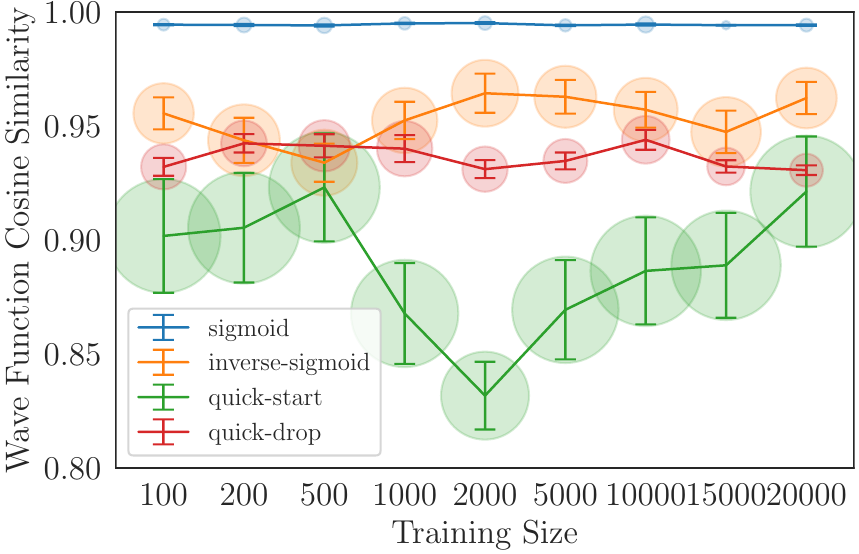}
    \hfill
    \includegraphics[width=0.44\textwidth]{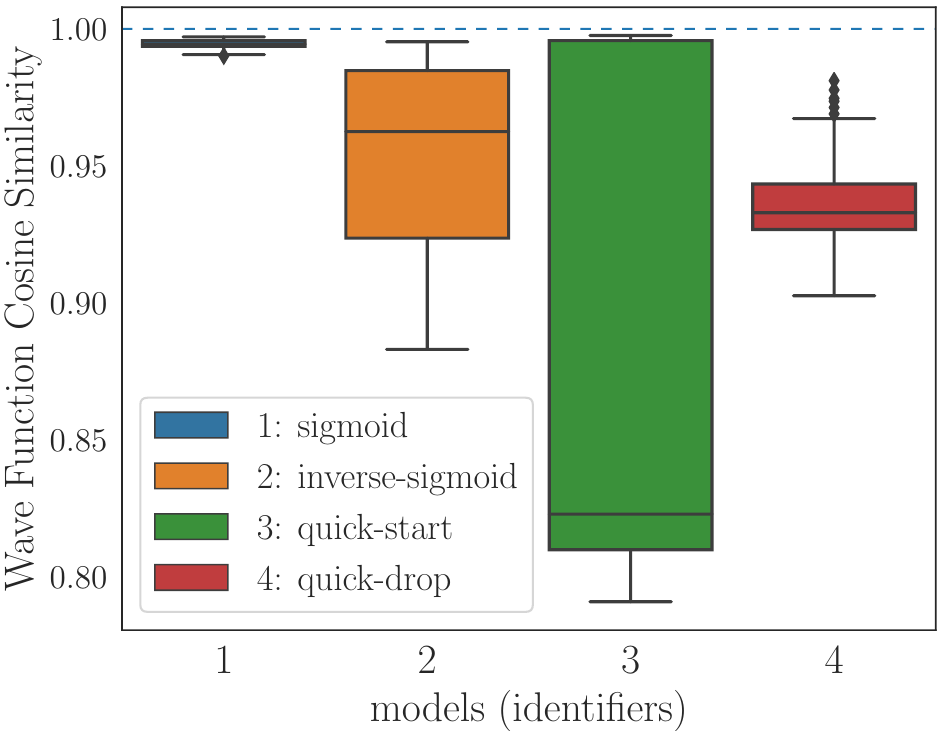}
    \caption{Wave function cosine similarity of different adaptive $C$-loss w.r.t. different training size. Left: cosine similarity on different training sizes. Right: Mean cosine similarity over all training sizes.}
    \label{fig:hyperparam}
\end{figure}

Using the hyper-parameter values to set up models, and run 10 times per setting on different training sizes, the results are shown in Fig. \ref{fig:hyperparam}. We can see that $sigmoid$ consistently performed better than other modes in both stability and accuracy. Another important information it conveys is that, \emph{quick-start}, like $sigmoid$ though it increases the weight of \nsloss{} much faster, results in a much more unstable results. Actually our results show that $quick-start$ dominates the leaderboard of both top-10 best and worst performances (also showed in the barplot in Figure \ref{fig:hyperparam}). It implies that a smooth and gradual switch of dominance between different loss terms is better in terms of stability.

\subsection{Electromagnetic Propagation Application}
\label{sec:em-appendix}
For this application, due to the prohibitively expensive training time, we did not conduct an exhaustive hyper-parameter search for \proposed{}. Instead, we manually tweaked the hyper-parameters until the optimization converged to an acceptable solution on the training sets $\mathcal{D}_{Tr}$ and $\mathcal{D}_{U}$ (note that the test labels were not used in any way during the hyper-parameter search). The final setting of hyper-parameters that we used are $\lambda_{C0}=0, \lambda_{S0}=1, \alpha_S=0.95, \alpha_C=0.01, T=50, \text{and } T_a=2000$. In terms of architecture, we used a fully-connected ANN, with 2 hidden layers of width 100.  We fed the $10$ permittivity values as input to the neural network, which produced a $802$-length vector (comprising of real and imaginary values) on the output layer. We trained the model for about 3000 epochs, and selected the best snapshot based on the lowest training error.



In order to account for the presence of complex values in the computation of label loss, we used an overlap-integral-based loss instead of $\text{Train-MSE}$ to measure the predictive accuracy on the labeled set $\mathcal{D}_{Tr}$. In particular, we replaced $\text{Train-MSE}$ with $\text{Train-Overlap}$ that is defined as:

\begin{equation}
    \text{Train-Overlap}:=\frac{1}{N} \sum_i \left( -||\frac{\hat{\eigenvec_i}}{||\hat{\eigenvec_i}||}\cdot\frac{\eigenvec_i^*}{||\eigenvec_i^*||}||^2 + ||\hat{\eigenval}_i - \eigenval_i||^2 \right)
\end{equation}

Minimizing $\text{Train-Overlap}$ ensures that the eigenvalues are correct as well as the profiles of the predicted eigenvector fit that of the ground-truth on the labeled set.

\printunsrtglossary[type=symbols,style=long,title={Table of Hyper-parameters used in \proposed{}}]













\end{document}